\documentclass[runningheads]{llncs}

\usepackage{eccv}

\usepackage{eccvabbrv}

\usepackage{graphicx}
\usepackage{booktabs}
\usepackage{float}
\usepackage{subcaption}
\usepackage{enumitem}

\usepackage[accsupp]{axessibility}  

\usepackage{hyperref}

\usepackage{orcidlink}

\begin{document}

\title{ShellMaker: Language-Guided Exterior Completion under Structural Constraints} 


\author{Ruiqi Xu\inst{1}\orcidlink{0009-0001-7874-0725} \and
Daniel Aliaga\inst{1}\orcidlink{0000-0001-9794-462X}}

\authorrunning{R. Xu \& D. Aliaga}

\institute{Purdue University, Lafayette IN 47907, USA \\
\email{\{xu2201,aliaga\}@purdue.edu}}

\maketitle

\begin{figure}[t]
  \centering
  \includegraphics[width=\textwidth]{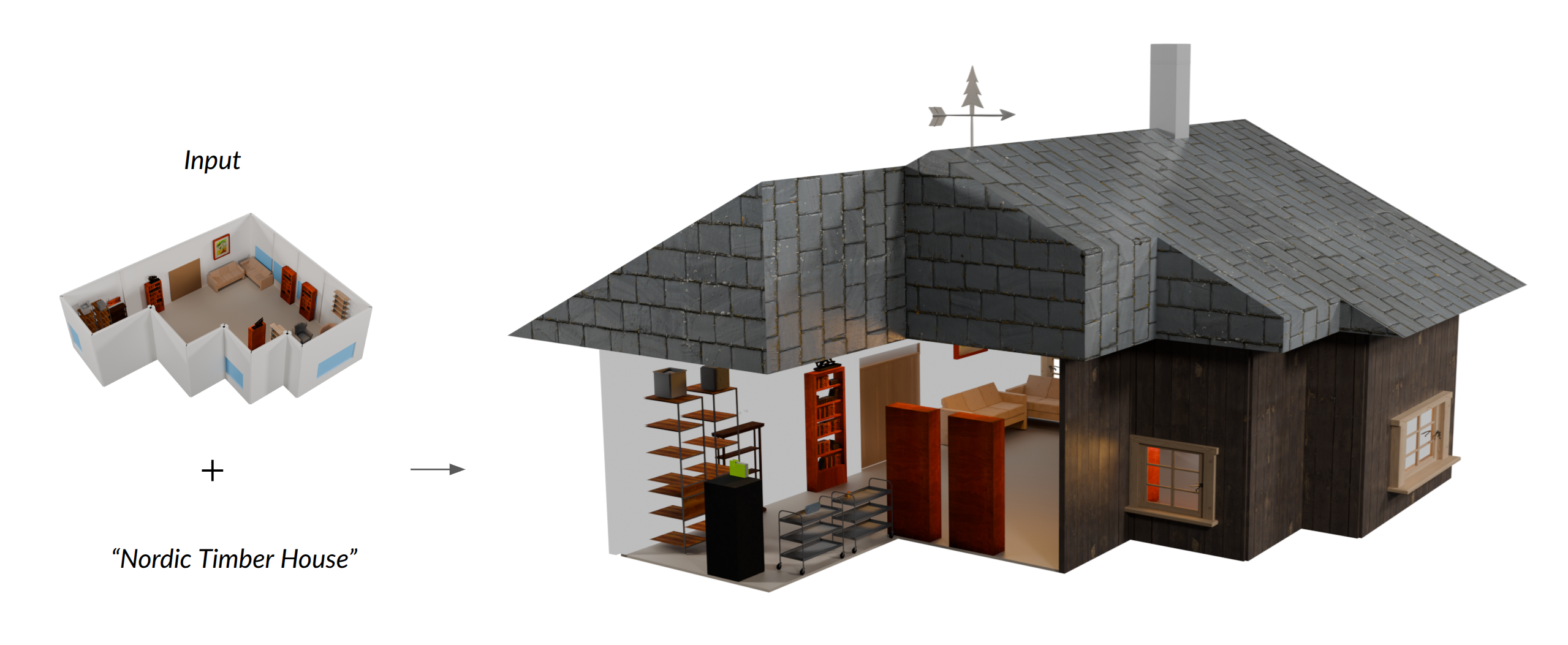}
  \caption{ShellMaker. Given a building scaffold and a style text prompt, our system produces a complete, textured 3D exterior. The wall geometry and window/door opening placements are preserved exactly across all style variations.}
  \label{fig:teaser}
\end{figure}

\begin{abstract}

Despite advances in indoor scene generation, synthesizing coherent building exteriors consistent with generated interiors remains largely unexplored. Existing methods can generate floor plans and wall layouts but typically stop at a structural shell, lacking stylistically consistent facades and roofs. Completing these exteriors is challenging because the footprint, wall geometry, and opening semantics must remain fixed—constraints that unconstrained generative models often violate. We introduce ShellMaker, a language-guided exterior completion framework that operates under these structural constraints. Given a building scaffold and a text style prompt, ShellMaker generates a complete exterior mesh with PBR materials by combining parametric roof generation, LLM-based part-aware prompt refinement, joint wall–roof material retrieval, and geometry-aware assembly. Operating on a format-agnostic scaffold representation, ShellMaker generalizes to indoor generators, CityGML, and CAD inputs, while maintaining structural consistency and improving architectural coherence over retrieval and unconstrained generative baselines. The project page is available at \url{https://ruiqixu37.github.io/ShellMaker_web/}.

  \keywords{Scene generation, Urban modeling, LLMs, Procedural Modeling}
\end{abstract}

\section{Introduction}
\label{sec:intro}

Recent indoor scene generation systems have made remarkable progress in automated building content creation. Modern pipelines can synthesize plausible interiors with structured floor plans, consistent wall massing, and explicit door and window openings \cite{yang2024holodeck, yang2025sceneweaver, wang2020sceneformer, gu2025artiscene, infinigen2024indoors}. However, these systems stop at the building shell, leaving facades, roofs, and stylistic exterior details unspecified. The resulting buildings are therefore incomplete and unsuitable for downstream applications such as digital twins, architectural visualization, or urban modeling.

Exterior completion requires generated assets to strictly follow the fixed footprint and opening semantics, as they will otherwise not align with the interior layout. Although recent advances in controllable 3D generation have improved conditional synthesis \cite{dong2024coin3d, fedele2025spacecontrol, jin2025sat, huang2025buildingblock}, these models cannot be reliably applied to exterior completion. This limitation stems from building data scarcity, as large-scale paired datasets linking structured building layouts to fully textured exterior meshes are not currently available. Consequently, a gap remains between perceptually realistic generation and layout-faithful exterior synthesis.

We address this problem with ShellMaker, a language-guided framework for exterior completion under hard structural constraints. Given a style prompt and a building scaffold specifying wall massing and opening layouts, ShellMaker generates a stylistically coherent exterior while strictly preserving the building footprint and layout semantics with three coordinated stages. First, a structural stage infers the fixed footprint from the scaffold and procedurally generates parametric roof geometry subject to the footprint constraint. Second, a style stage retrieves compatibility-aware wall–roof material pairs and uses a large language model (LLM) to translate the user prompt into part-aware specifications, which are used by pretrained 3D generative models to synthesize detailed, style-consistent architectural elements at the part level. Third, an assembly stage performs geometry-aware opening carving, semantics-aligned asset placement, and scale-consistent UV parameterization, ensuring precise alignment with predefined openings and coherent exterior composition.

Although motivated by indoor scene generators, ShellMaker operates on a format-agnostic scaffold abstraction. This enables, in addition, direct application to CityGML building shells and CAD-derived floor plans without modifying the pipeline. We define quantitative structural consistency metrics that measure footprint violations and layout alignment. Altogether and across diverse scaffolds and style prompts, ShellMaker improves architectural coherence, material harmony, and constraint satisfaction as compared to independent retrieval pipelines and unconstrained generative baselines. Our results and analysis include buildings based on over 200 diverse scaffolds, 500+ textures, and 60 style prompts. Further, ShellMaker integrates seamlessly with indoor scene generation pipelines, enabling the automatic synthesis of complete building assets with coherent interiors and structurally consistent exteriors.

\noindent To summarize ShellMaker's contributions:
\begin{enumerate}
    \item We formulate the task of language-guided exterior completion under hard scaffold constraints, turning partial building scaffolds into complete, style-consistent assets.
    \item We propose a modular pipeline to integrate procedural structure generation, part-level stylization, and compatibility-aware wall--roof material retrieval that limits cross-stage error propagation and enables asset reuse.
    \item We introduce a geometry-aware assembly process that enforces footprint, wall-boundary, and opening preservation, yielding editable, part-structured exterior meshes.
\end{enumerate}

\section{Related Works}

\paragraph{3D Generative Models.}
Recent advances in text-to-3D and image-to-3D generation have enabled the synthesis of detailed 3D assets from natural language or visual inputs \cite{xiang2025structured, chen2025sam, lin2025kiss3dgen, wu2025direct3d, ye2025hi3dgen}. Feed-forward 3D diffusion models and reconstruction pipelines can produce textured meshes with strong appearance fidelity, with some methods additionally predicting physically-based-rendering (PBR) material parameters such as diffuse, normal, and roughness maps \cite{xiang2025native, hunyuan3d2025hunyuan3d, seed2025seed3d}. However, these methods are typically designed for unconstrained generation and do not enforce predefined geometric layouts. As a result, generated geometry often deviates from required structural constraints when applied to structured architectural inputs.

\paragraph{Controllable 3D Generation.}
Several works introduce controllability through spatial conditions such as depth and normal maps \cite{xie2023sparsefusion, qiu2024richdreamer}, part-level semantics \cite{lin2025partcrafter, yang2025holopart}, or geometric priors \cite{dong2024coin3d, jin2025sat, fedele2025spacecontrol}. These approaches improve alignment between generated geometry and input signals, but the control remains approximate and does not guarantee strict preservation of topology or layout semantics. In particular, existing methods lack mechanisms to jointly enforce footprint consistency while maintaining semantic alignment with text prompts. In contrast, our method operates under hard structural constraints, treating the building footprint and opening placements as immutable while generating stylistically consistent exterior components.

\paragraph{Indoor Scene and Building Generation.}
Recent indoor scene generation systems can synthesize structured interiors with semantically grounded floor plans, wall layouts, and explicit door and window openings \cite{yang2024holodeck, yang2025sceneweaver, wang2020sceneformer, gumin2025procedural, bokhovkin2024scenefactor, tam2025sceneeval, gu2025artiscene, infinigen2024indoors}. However, these pipelines typically stop at the structural shell and do not produce coherent building exteriors. Recent works attempt exterior generation conditioned on coarse geometric priors such as 3D shape hints or semantic maps \cite{jin2025sat, huang2025buildingblock, chang2021building}, but they do not fully preserve the input footprint, lack mechanisms to jointly enforce structural constraints and semantic style, and are not designed to operate directly on structured outputs from indoor scene generators. Earlier work in procedural urban modeling generates buildings from rule-based grammars but requires manual specification and lacks language-driven style control \cite{nishida2018procedural, muller2006procedural, vanegas2010building}. In contrast, ShellMaker operates directly on structured building scaffolds produced by indoor generators, CAD models, or CityGML data, enabling the automatic completion of layout-consistent exteriors and the generation of complete building assets.


\section{Method}
\label{sec:method}
\begin{figure}[t]
    \centering
    \includegraphics[width=\linewidth]{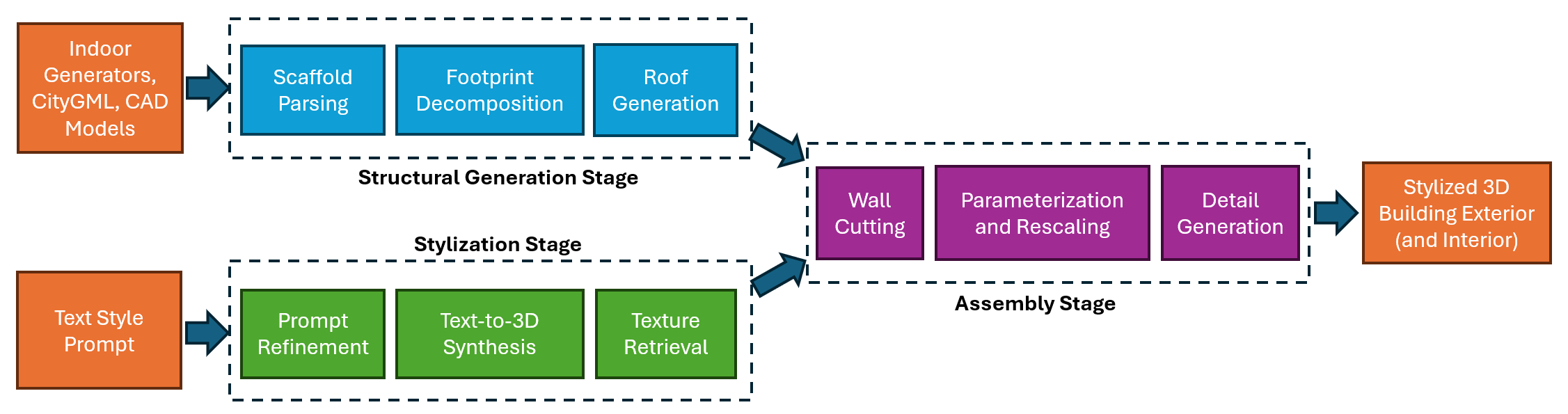}
    \caption{\textbf{ShellMaker Pipeline}. Given a structured building scaffold (e.g., indoor scene generators, CityGML, or CAD models) and a text style prompt, ShellMaker synthesizes a complete stylized 3D exterior through three stages. The structural stage parses the scaffold and generates parametric roof geometry, while the style stage refines the prompt using an LLM, retrieves compatible wall–roof materials, and generates style-conditioned architectural parts. These components are combined through structure-aware assembly stage, producing a layout-preserving, style-consistent building exterior.}
    \label{fig:overview}
\end{figure}

\subsection{Problem Formulation}
Given a structured building scaffold $\mathcal{S}$ and a text style prompt $\mathcal{P}$, our goal is to generate a textured 3D exterior mesh
\begin{equation}
    \mathcal{E} = \Phi(\mathcal{S}, \mathcal{P}),
\end{equation}
that is both geometrically faithful to the scaffold and semantically aligned with the requested architectural style. The scaffold $\mathcal{S}$ encodes the building footprint, wall boundaries, and the spatial placement of door and window openings, and may be sourced from structured scene graphs produced by deep-learning based indoor scene generators, CAD models, or CityGML files. The output $\mathcal{E}$ is a fully assembled exterior mesh with PBR-ready material parameters, consisting of textured wall surfaces, procedurally generated roof geometry, style-consistent door and window assets inserted into predefined openings, and optional roof ornaments.

To ensure compatibility with indoor scene generators and structured building pipelines, the synthesized exterior must satisfy strict structural consistency. In particular, the footprint topology, wall boundaries, and opening placements specified by the scaffold need to be preserved exactly. Beyond geometry, the exterior should also reflect the semantic style specified by the user prompt.

As illustrated in (Fig.~\ref{fig:overview}), ShellMaker decomposes the synthesis
process into three coordinated stages:
\begin{equation}
    \Phi = \Phi_{\text{assemble}} \circ
           \bigl(\Phi_{\text{structure}},\;\Phi_{\text{style}}\bigr),
\end{equation}
where $\Phi_{\text{structure}}(\mathcal{S})$ constructs the geometric scaffold and generates parametric roof geometry from the fixed footprint (Sec.~\ref{para:structual_stage}), $\Phi_{\text{style}}(\mathcal{P})$ translates the user prompt into part-aware specifications using a large language model, synthesizes style-consistent door and window assets with a pretrained 3D generative model, and retrieves compatible wall--roof material pairs (Sec.~\ref{para:style_stage}), and $\Phi_{\text{assemble}}$ combines the two stages to produce the final exterior mesh through geometry-aware opening carving, semantics-aligned asset placement, and scale-consistent UV parameterization (Sec.~\ref{para:assmbly_stage}). 

\subsection{Structure Generation Stage}
\label{para:structual_stage}

\paragraph{Scaffold Parsing.}
The first stage converts heterogeneous architectural inputs into a unified \emph{scaffold} representation that encodes walls, floors, ceilings, and door/window openings in a common schema. We support three classes of inputs:
\begin{enumerate} 
    \item {\emph {Indoor Scene Generators.} The parser extracts wall segments, floor polygons, and door/window opening metadata from the scene graph.}
    \item {\emph {CityGML Models.} The system parses semantically tagged surfaces (e.g., \texttt{WallSurface}, \texttt{RoofSurface}, \texttt{GroundSurface}) and merges coplanar wall fragments that belong to the same façade plane. Openings are projected into each wall's local frame to recover their width, height, and placement.}
    \item {\emph {BIM/CAD Data.} Geometric primitives are tessellated into triangle meshes using \texttt{ifcopenshell}. Wall footprints and normals are inferred from mesh geometry, and opening metadata is recovered either from explicit Industry Foundation Classes (IFC) attributes or from mesh-based bounding boxes.}
\end{enumerate}

All parsers produce a normalized scaffold representation containing wall segments with 2D endpoints and outward normals, floor and ceiling surfaces, opening placements in both world and wall-local coordinates, and floor-plan polygons for downstream roof generation. This unified scaffold enables the subsequent structural, style, and assembly stages to operate regardless of the input format.

\paragraph{Footprint Decomposition.}
To generate roofs for complex footprints, we first partition the building footprint into simpler subregions. ShellMaker supports two user-controlled decomposition strategies. The first strategy is straight-skeleton decomposition, which computes the medial axis of the floor-plan polygon and partitions the footprint into cycle regions associated with skeleton ridges. Adjacent regions may be merged under a convexity tolerance ($\epsilon{=}0.02$) to reduce over-fragmentation. This representation preserves the geometric structure of the original footprint and naturally supports conformal roof generation for non-convex layouts. 

The second strategy enables a wider range of roof compositions by providing an alternative greedy rectangle decomposition that recursively extracts the largest axis-aligned rectangle contained in the footprint, subtracts it, and repeats on the residual polygons until the remaining area falls below a threshold. The resulting rectangular subregions are then assigned independent roof primitives, enabling diverse multi-part roof structures over complex footprints.

\begin{figure}[tb]
  \centering
  \begin{subfigure}{\linewidth}
    \centering
    \includegraphics[width=0.8\linewidth]{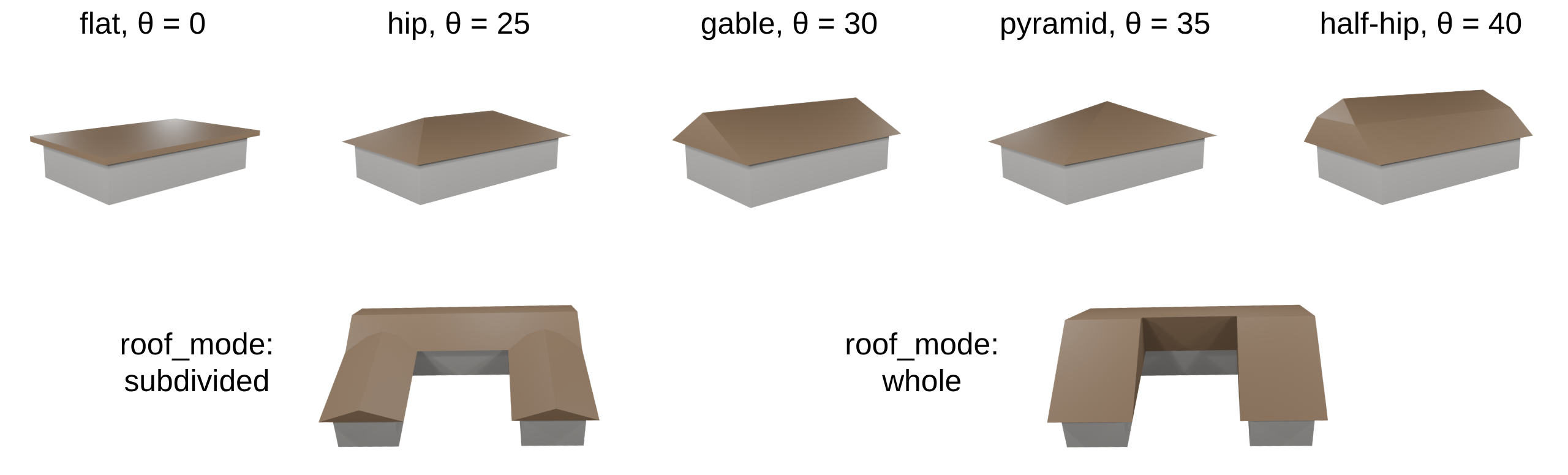}
  \end{subfigure}
  \caption{Parametric Roof Generation. Top: canonical roof types supported by ShellMaker (flat, hip, gable, pyramid, and half-hip) with example pitch angles $\theta$. Bottom: two roof generation modes for complex footprints.}
  \label{fig:short}
\end{figure}

\paragraph{Roof Generation.}
Given a footprint or set of subregions, ShellMaker generates parametric roof geometry from five canonical roof types: \emph{flat}, \emph{gable}, \emph{pyramid}, \emph{hip}, and \emph{half-hip}. Roof geometry
is additionally controlled through pitch angle ($\theta_{\text{pitch}}$), eave overhang ($d_{\text{overhang}}$), ridge-length fraction ($f_{\text{ridge}}$), and half-hip clipping fraction ($f_{\text{hip}}$), enabling continuous variation in roof profiles.

Like the decomposition strategy, the roof construction mode is a user-selectable option. ShellMaker supports two roof construction modes. In \emph{whole} mode, a single roof is generated over the entire footprint. In \emph{subdivided} mode, roofs are generated independently for each footprint subregion obtained from the decomposition stage and merged into a single mesh, enabling multi-part roof compositions over complex layouts.

\subsection{Stylization Stage}
\label{para:style_stage}

\paragraph{Part-aware Prompt Refinement with LLM.}
As ShellMaker is designed to support free-form text style prompts, inputs such as ``a Victorian bakery'' describe the building only at a coarse semantic level and provide insufficient guidance for generating detailed architectural parts. Such descriptions rarely specify the geometry or decorative elements of individual components, leading to stylistically inconsistent results when used directly as text-to-3D prompts. Generating high-quality doors, windows, and roof ornaments therefore requires more explicit part-aware prompts that capture stylistic cues at the element level.

To bridge this gap, we employ an instruction-tuned LLM to automatically refine the input free-form prompt into a set of structured text-to-3D prompts. The system instruction enforces that each prompt describes a single isolated architectural element with no surrounding walls or environmental context, ensuring that the generated meshes can be composited cleanly onto the scaffold. The LLM is further guided to include style-authentic geometric and decorative cues appropriate to the architectural style while respecting size constraints for each variant. This refinement process is applied independently to the wall, roof, window, door, and roof-ornament components.

\begin{figure}[tb]
  \centering
  \begin{subfigure}{\linewidth}
    \centering
    \includegraphics[width=0.8\linewidth]{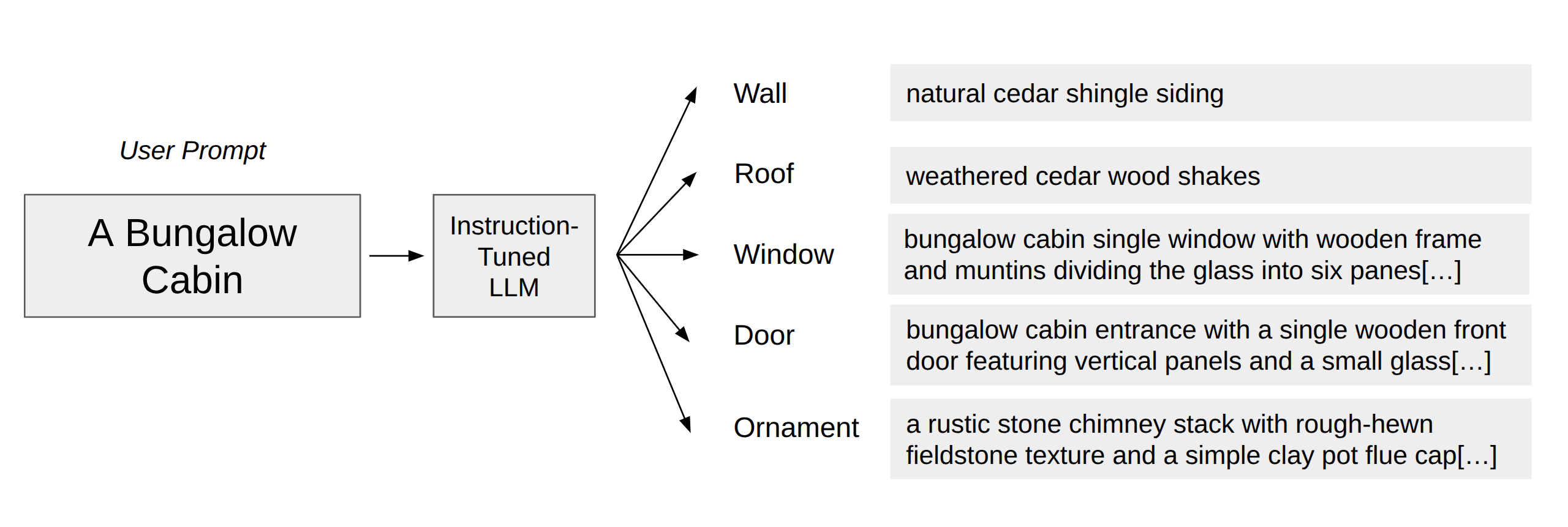}
    \label{fig:short-b}
  \end{subfigure}
  \caption{Part-aware Prompt Refinement. A free-form user prompt is passed to an instruction-tuned LLM, which generates a structured, style-consistent text prompt for each architectural component.}
  \label{fig:short}
\end{figure}

\paragraph{Two-stage Text-to-3D Synthesis.}
Each refined prompt results in a textured mesh by means of a two-stage pipeline: Nano Banana~\cite{team2023gemini} generates a reference image from the prompt, and Trellis-2~\cite{xiang2025native} reconstructs a textured mesh with PBR material parameters from the image. This factorization leverages the strong stylistic fidelity of modern image generators while delegating geometric reconstruction to a dedicated 3D backbone, enabling style-consistent asset synthesis without per-style supervision or fine-tuning.

\paragraph{Joint Wall-Roof Texture Retrieval.}
Rather than generating surface materials from scratch, we adopt a retrieval-based approach for wall and roof textures. Current generative texture models often struggle to produce tileable, physically consistent PBR materials with aligned diffuse, normal, and roughness maps -- properties essential for large architectural surfaces. Unfortunately, naive per-surface retrieval can yield perceptually inconsistent combinations, because facade and roof materials exhibit strong cross-surface dependencies in real-world architecture. To address this issue, we introduce a compatibility-aware joint retrieval mechanism that precomputes cross-surface compatibility offline and incorporates it during runtime texture selection.

\paragraph{Offline Compatibility Estimation.}
Let $\mathcal{W}{=}\{w_i\}_{i=1}^{N_w}$ and $\mathcal{R}{=}\{r_j\}_{j=1}^{N_r}$ be PBR texture-pack libraries for walls and roofs (each containing diffuse, normal, and roughness maps).  We precompute a compatibility matrix $C\in[0,1]^{N_w\times N_r}$ whose entries combine four signals:
\begin{equation}
  C_{ij} \;=\;
    \lambda_1 C^{\text{mat}}_{ij} +
    \lambda_2 C^{\text{color}}_{ij} +
    \lambda_3 C^{\text{freq}}_{ij} +
    \lambda_4 C^{\text{clip}}_{ij},
  \qquad
  \lambda = (0.45,\, 0.25,\, 0.15,\, 0.15).
\end{equation}
$C^{\text{mat}}$ assigns high scores to commonly co-occurring architectural pairings (e.g.\ brick walls with clay-tile roofs, concrete facades with metal roofs). $C^{\text{color}}$ measures color harmony via mean $L^*a^*b^*$ distance and saturation difference of the diffuse maps.  $C^{\text{freq}}$ compares radial power spectra (2D FFT) by cosine similarity, capturing texture-scale alignment. Finally, $C^{\text{clip}}$ computes CLIP embedding similarity between diffuse maps as a high-level semantic signal. These scores are computed once offline for all texture pairs.

\begin{figure}[tb]
  \centering
  \begin{subfigure}{\linewidth}
    \centering
    \includegraphics[width=1.0\linewidth]{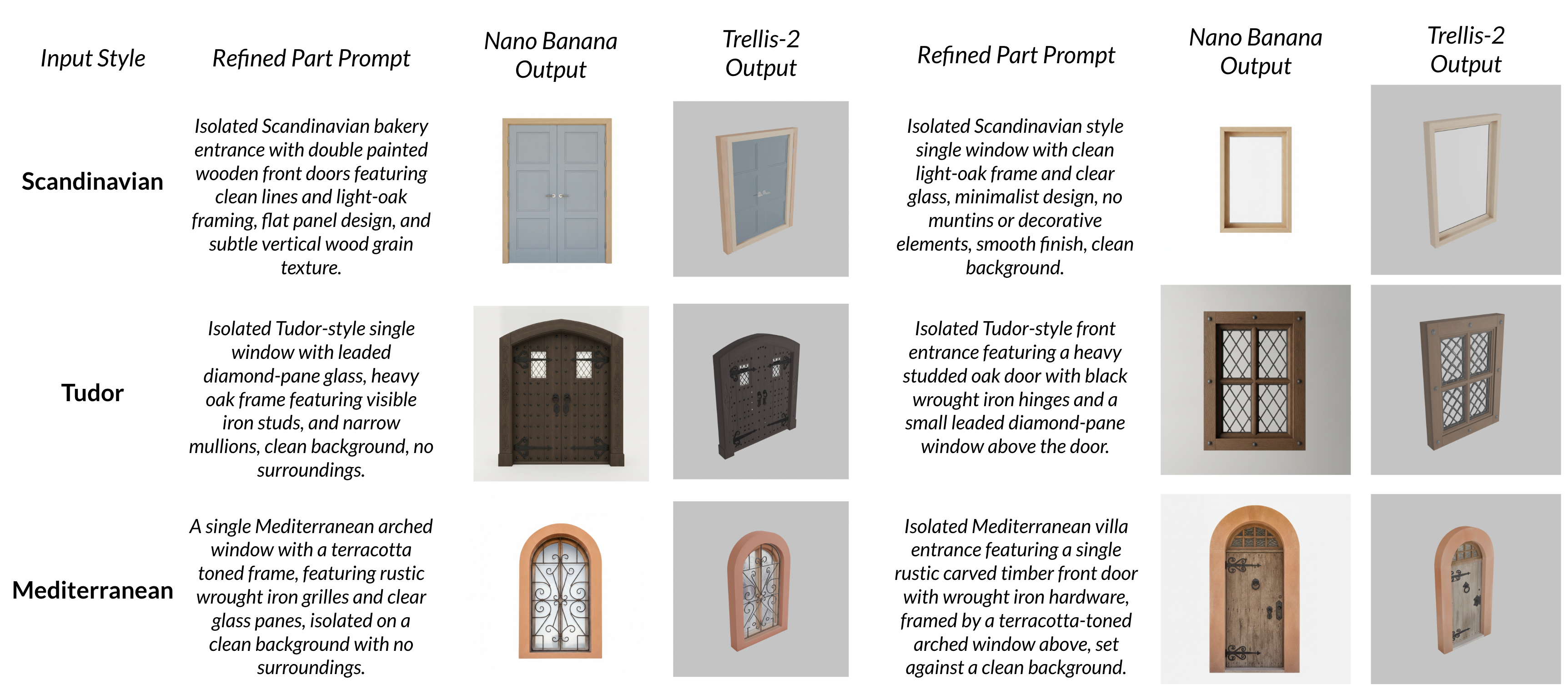}
    \label{fig:short-b}
  \end{subfigure}
  \caption{Two-stage part synthesis intermediates. LLM-refined part prompts, Nano Banana reference images, and reconstructed Trellis-2 meshes across three styles and two element types. Style-specific geometric and decorative cues introduced during prompt refinement carry through to the final textured parts.}
  \label{fig:short}
\end{figure}

\paragraph{Runtime Joint Retrieval.} At runtime, Shellmaker computes the CLIP similarities of the refined wall and roof prompts $t_w$ and $t_r$ against the texture libraries: 
$$s_w(i)=\cos(CLIP(t_w),CLIP(w_i)), s_r(j)=\cos(CLIP(t_r),CLIP(r_j))$$ 

We first shortlist the top-$K$ wall textures according to $s_w(i)$ and the top-$K$ roof textures according to $s_r(j)$, producing $K^2$
candidate pairs. Each pair is scored using the precomputed compatibility matrix:
\begin{equation}
  s_{ij} =
    \alpha\, s_w(i) +
    \beta\, s_r(j) +
    \gamma\, C_{ij},
  \qquad
  (\alpha,\beta,\gamma) = (0.40,\, 0.30,\, 0.30).
\end{equation}
To avoid deterministic selections and encourage stylistic variation, we retain the top-$M$ pairs and sample the final pair using a temperature-controlled softmax $p_{ij}\propto\exp(s_{ij}/\tau)$. Lower $\tau$ favors conservative combinations while higher $\tau$ encourages diversity.

\subsection{Assembly Stage}
\label{para:assmbly_stage}
The final stage assembles the scaffold, roof geometry, generated assets, textures, and ornaments into a single mesh, resolving three geometry-aware sub-problems: UV parameterization, opening cutting with asset insertion, and optional architectural detailing.

\paragraph{Shape-Conforming Wall Cutting.}
For each inserted part, the pipeline computes a 2D boundary polygon that drives the wall opening cut. It first determines the part's body bounds by trimming the outermost $5\%$ of surface area along each axis, excluding ornamental protrusions (e.g., pediments, pilasters, platforms) that inflate the bounding box. The mesh is then clipped at its back-plane to remove interior-protruding geometry. If the resulting silhouette is sufficiently rectangular (i.e., projected area $\geq 95\%$ of its bounding box), a fast rectangular cut is used directly. Otherwise, the mesh is scaled to match the target opening dimensions and a 2D boundary polygon is extracted via a binned silhouette envelope. The X-span is divided into 64 bins and the upper and lower Y-extents per bin are traced into a closed polygon, simplified, and outward-buffered by 0.01 to ensure full coverage. The polygon is finally rescaled to fit exactly within the target opening dimensions before being used as the wall cutter through boolean subtraction.

\paragraph{UV Parametrization and Auto-rescaling.}
Each wall and roof face is assigned world-space planar UVs by projecting onto the plane perpendicular to the dominant component of its unit normal. Roof meshes use a dedicated \emph{skirt mode}, where horizontal and sloped roof faces use a ground-plane projection, while vertical fascia faces anchor their UVs to the same horizontal coordinates with a vertical offset measured from the roof edge. This prevents visible seams along the eave-slope boundary.

To maintain consistent perceptual texture density across different materials, we apply an automatic UV rescaling factor precomputed offline for each texture. For every diffuse texture map, we estimate its dominant spatial frequency by computing a 2D FFT of the luminance image and performing radial averaging of the power spectrum to obtain a 1D frequency profile. The power-weighted mean frequency (excluding very low-frequency components) characterizes the typical feature scale of the material. Across a texture library, the median dominant frequency is used as a reference; each texture receives a scale factor $s=\text{median}/f_{\text{dominant}}$. Coarse materials (e.g., large brick patterns) therefore repeat more frequently across a surface, producing smaller visible tiles, while fine-grained materials repeat less frequently. At runtime this factor multiplies user-specified UV scale to produce the final tiling frequency used in planar UV assignment, ensuring consistent perceptual material scale across facades and roofs.

\paragraph{Optional Architectural Details.}
ShellMaker optionally augments the base structure with procedural roof utilities and decorative ornaments. Downspouts are generated through a gutter-to-downspout pipeline that first extracts eave edges from the roof mesh and chains them into continuous gutter polylines. In addition, roof ornaments are generated from a small set of architectural categories (e.g., chimneys, finials, and dormers). For each placement, the ornament is embedded into the roof surface, and clipped against the roof geometry so that only geometry above the roof remains. This clipping step ensures ornaments conform to sloped roofs while preventing intersections with underlying roof panels.

\section{Experiments}
\subsection{Datasets}

\begin{figure*}[t]
\centering
\includegraphics[width=0.8\textwidth]{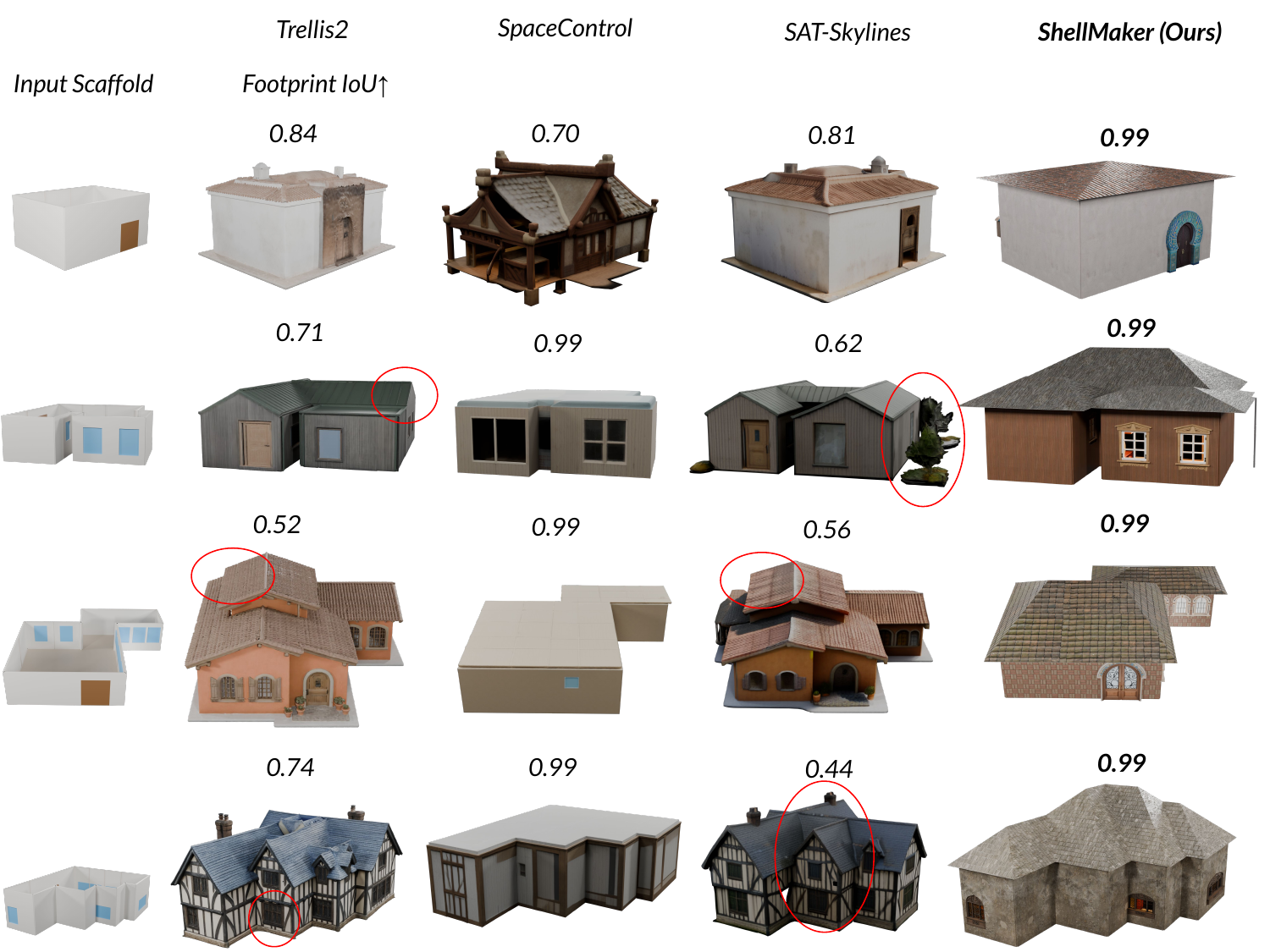}
\caption{Given the same scaffold and style prompt, generative baselines often produce visually plausible exteriors but fail to faithfully preserve the scaffold footprint, exhibiting incorrect aspect ratios, omitted footprint boundaries, or geometry extending beyond the scaffold. Methods that better respect the footprint frequently do so at the cost of degraded texture quality. In contrast, ShellMaker preserves the scaffold geometry while producing stylistically consistent exteriors. Red circles highlight subtle footprint mismatches, objects outside the scaffold, and incorrect structural semantics.}
\label{fig:baseline-comparison}
\end{figure*}

\label{sec:exp_data_prompts}
\paragraph{Building Data.} Our primary evaluation uses 100 scaffolds exported from indoor scene generation pipelines, spanning 10 building types (e.g., apartment, bookstore, bakery, gym, \ldots) with diverse footprint geometries and opening configurations. To test format generality, we additionally source 50 CityGML-derived building shells and 50 CAD-derived models from the Internet, for a total of 200 test scaffolds. All sources are converted into our unified representation.

\paragraph{Wall and Roof Textures.} We construct an asset pool of 498 wall textures and 81 roof textures by manually filtering assets from CC0 Textures and Poly Haven. All textures are $4$k resolution, include diffuse, normal, and roughness maps, and are released with a free license, enabling PBR of generated buildings.

\paragraph{Prompt Suite.} We define 60 prompt templates across three categories:
\emph{style-only} (20 templates specifying architectural character without material constraints; e.g., \emph{"a Victorian bakery"}), \emph{material-only} (20 templates specifying wall and roof finishes without a style descriptor; e.g., \emph{"a building with red brick walls and a grey slate roof"}), and \emph{mixed} (20 templates combining a style descriptor with explicit material clauses; e.g., \emph{"Tudor style with red brick walls"}). The full suite is provided in the supplementary material.

\subsection{Implementation Details}
Prompt refinement and ornament planning use GPT-4.1 with structured JSON output. Reference images are generated with Nano Banana and lifted to textured meshes by Trellis-2. Generated assets are cached per unique <style, category, part-name> tuple, so openings sharing the same canonical part name in the scaffold metadata reuse a single mesh for regularity. For evaluation, all methods are rendered under identical conditions, with 8 azimuth views at $30^{\circ}$ elevation in a turntable configuration plus 2 street-level perspective views aligned to principal facades, lit by a fixed neutral High Dynamic Range Imaging (HDRI) environment map with constant exposure. All code and data will be released.

\begin{table}[t]
\centering
\caption{Main quantitative evaluation on style alignment and structural consistency.}
\label{tab:main}
\setlength{\tabcolsep}{4pt}
\small
\begin{tabular}{lccccc}
\toprule
 & \multicolumn{2}{c}{Style Alignment} & \multicolumn{3}{c}{Structural Consistency} \\
\cmidrule(lr){2-3} \cmidrule(lr){4-6}
Method
& CLIP-Sim$\uparrow$
& UNI3D$\uparrow$
& Footprt IoU$\uparrow$
& Opening IoU$\uparrow$
& Center L2$\downarrow$ \\
\midrule
Trellis-2                    & \textbf{0.318} & 0.401 & 0.492 & 0.431 & 0.914 \\
SpaceControl               & 0.274          & 0.258 & 0.812 & 0.508 & 0.523 \\
SAT-Skylines               & 0.281          & 0.265 & 0.761 & 0.548 & 0.463 \\
\midrule
\textbf{ShellMaker}        & 0.315          & \textbf{0.424} & \textbf{0.992} & 0.883 & 0.091 \\
\quad- Joint Texture       & 0.285          & 0.357 & 0.992 & 0.889 & \textbf{0.083} \\
\quad- LLM Prompt          & 0.293          & 0.364 & 0.992 & \textbf{0.922} & 0.095 \\
\quad- Both                & 0.280          & 0.305 & 0.992 & 0.881 & 0.084 \\
\bottomrule
\end{tabular}
\end{table}

\subsection{Main Quantitative Results}
We compare ShellMaker with recent 3D generative baselines including Trellis-2 \cite{xiang2025native}, SpaceControl\cite{fedele2025spacecontrol}, and SAT-Skylines\cite{jin2025sat}. To provide strong image-based baselines for Trellis-2 and SAT-Skylines, we attempt to align generation with the input scaffold using image-level guidance. We render depth and semantic maps from the scaffold, and prompt Nano Banana to generate a plausible building image conditioned on these signals. The resulting image is then used as the input to Trellis-2 and SAT-Skylines. SAT-Skylines, along with SpaceControl, are methods that extend the Trellis-1 framework by injecting geometric shape priors into the pretrained latent space to encourage consistency with the input structure during generation.

Table~\ref{tab:main} reports both style alignment and structural consistency metrics across all methods. Style alignment is measured using CLIP-Sim, computed as the average multi-view CLIP similarity between rendered images and the input prompt, and UNI3D, which provides complementary view-independent evaluation. Structural fidelity is evaluated using footprint IoU, opening alignment IoU, and the average L2 distance between predicted and target opening centers. Opening alignment IoU measures overlap between predicted and scaffold opening regions, capturing agreement in size and extent, but it can be overly strict when the generator produces non-rectilinear openings (e.g., arched windows) that naturally deviate from the scaffold’s rectangular semantic box, without being structurally incorrect. We therefore also report the average L2 distance between predicted and target opening centers, which measures positional accuracy independent of shape/area.

Generative baselines achieve reasonable prompt alignment but often fail to satisfy structural constraints. Trellis-2 achieves the highest CLIP-Sim among all methods (0.318) but scores poorly on structural metrics, with a footprint IoU of only 0.492 and an opening center error of 0.914, confirming that image-conditioned 3D reconstruction cannot enforce hard geometric constraints. SpaceControl and SAT-Skylines improve structural fidelity by injecting geometric priors into the latent space, yet both still exhibit substantial opening misalignment and suffer notable drops in style alignment relative to Trellis-2.

In contrast, ShellMaker achieves strong performance across both style and structural metrics. It attains a footprint IoU of 0.992 and an opening center error of only 0.091, indicating near-perfect preservation of scaffold boundaries and opening placement, which makes the method well suited for integration with indoor scene pipelines. ShellMaker produces stylized architectural elements (e.g., arched windows or decorative doors), which may deviate from the rectangular semantic boxes provided by the scaffold; consequently, the default prompt ablation, which always generate simpler rectangular openings, can achieve slightly higher Opening IoU. However, when considered together with the center error metric, the results show that ShellMaker still maintains highly accurate opening placement. On style alignment, ShellMaker matches Trellis-2's CLIP-Sim while substantially outperforming all baselines on the view-independent UNI3D signal. Removing joint texture selection reduces CLIP-Sim from 0.315 to 0.285 and UNI3D from 0.424 to 0.357, while disabling LLM-based prompt refinement similarly degrades both style scores, with the largest drop observed when both are removed. Structural consistency remains stable across all ablations, and footprint IoU is identical across ShellMaker variants because the pipeline never alters the underlying wall massing. Roofs and facade elements are generated while the scaffold footprint and wall geometry remain fixed.

\begin{figure}[h!]
\centering
\includegraphics[width=\textwidth]{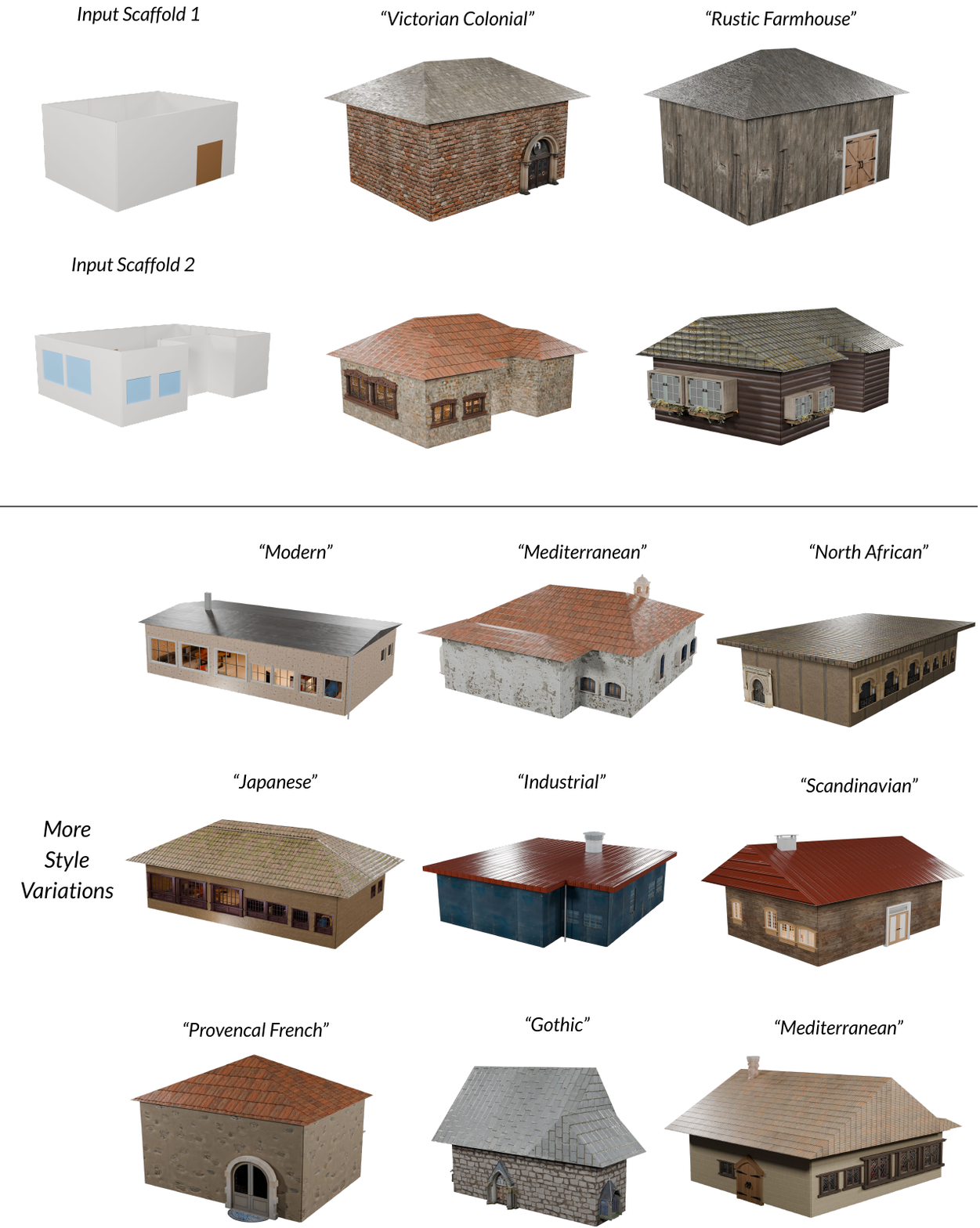}
\caption{\textbf{Style gallery.} (Top) Two input scaffolds each rendered under two style prompts, demonstrating preservation of wall geometry and opening placements across variants. (Bottom) Nine additional styles applied to various scaffolds, illustrating the breadth of architectural styles ShellMaker can automatically generate with appropriate roof forms, materials, and detailing.}
\label{fig:style-gallery}
\end{figure}

\subsection{Perceptual Evaluation}

We conduct a two-alternative forced choice (2AFC) evaluation using GPT-4.1 as the judge across 100 scaffold--prompt pairs. For each pair, the judge is presented with the input prompt and rendered views of ShellMaker alongside a random baseline, and asked to select the preferred output on overall quality, style fidelity, and structural plausibility. 

ShellMaker achieves win rates above 70\% overall against all three baselines (Table~\ref{tab:llm_eval}). Against Trellis-2, gains are most pronounced on structural plausibility (84.1\%), while style fidelity is more competitive (48.3\%), reflecting Trellis-2's stronger appearance modelling but weaker geometric grounding. Against SpaceControl and SAT-Skylines, ShellMaker leads consistently across all axes, with particularly large margins on structural plausibility (92.8\%, 91.2\%), confirming that ShellMaker produces outputs that better respect input scaffold constraints.

\begin{table}[t]
\centering
\caption{Additional evaluation results.}
\label{tab:additional_results}

\begin{subtable}[t]{0.5\linewidth}
\centering
\caption{Generalization across scaffold sources.}
\setlength{\tabcolsep}{3pt}
\small
\resizebox{\linewidth}{!}{
\begin{tabular}{lcccc}
\toprule
 & \multicolumn{2}{c}{Style} & \multicolumn{2}{c}{Structural} \\
\cmidrule(lr){2-3}\cmidrule(lr){4-5}
Source & CLIP-S$\uparrow$ & UNI3D$\uparrow$ & Ftpr.\ IoU$\uparrow$ & Open.\ L2$\downarrow$ \\
\midrule
Indoor Generator    & 0.315 & 0.424 & 0.992 & 0.091 \\
CityGML             & 0.308 & 0.418 & 0.991 & 0.104 \\
CAD                 & 0.301 & 0.409 & 0.889 & 0.118 \\
\bottomrule
\end{tabular}
}
\label{tab:generalization}
\end{subtable}
\hfill
\begin{subtable}[t]{0.45\linewidth}
\centering
\caption{LLM evaluation (2AFC, \%).}
\setlength{\tabcolsep}{3pt}
\small
\resizebox{\linewidth}{!}{
\begin{tabular}{lccc} 
\toprule
Method & Overall$\uparrow$ & Style$\uparrow$ & Struct.$\uparrow$ \\
\midrule
Trellis-2            & 71.2 & 48.3 & 84.1 \\
SpaceControl        & 80.1 & 71.2 & 92.8 \\
SAT-Skylines        & 87.4 & 68.8 & 91.2 \\

\bottomrule
\end{tabular}
}
\label{tab:llm_eval}
\end{subtable}

\end{table}


\subsection{Generalization Across Scaffold Sources}

Table~\ref{tab:generalization} evaluates ShellMaker on scaffolds derived from different structured building formats. Results on indoor generator inputs closely match the main evaluation, confirming that the system performs consistently on the scaffold representation produced by indoor scene generation pipelines. On CityGML inputs, style alignment decreases slightly while structural metrics remain nearly unchanged. Finally, CAD inputs exhibit noisier geometry and inconsistent topology, leading to a modest drop in footprint IoU and higher opening alignment error. Despite these challenges, ShellMaker maintains competitive style scores and preserves most structural constraints, indicating that the pipeline generalizes well across heterogeneous scaffold sources.

\subsection{Qualitative Results}
Figure~\ref{fig:style-gallery} presents representative outputs generated from indoor-scene scaffolds with different style prompts. The top examples show that ShellMaker preserves the input footprint and maintains accurate alignment of door and window openings while adapting roof geometry, materials, and architectural details to the requested style. The bottom gallery further demonstrates the stylistic range of the system across a variety of architectural themes, highlighting the ability of ShellMaker to produce diverse exteriors while respecting structural constraints. We provide more qualitative results and a fly-through video in the supplemental materials.

\subsection{Limitations}
ShellMaker's primary failure mode lies in the part generation pipeline, which can occasionally produce degenerate components such as malformed roof segments or geometrically inconsistent parts. In addition, part intersections may occur in semantically dense regions of the facade where multiple elements are placed in close proximity. We refer the reader to the supplementary material for illustrative failure cases.

ShellMaker currently treats the facade as a single massing shell and stylizes openings uniformly without modeling per-floor articulation. Richer multi-floor facade variation, such as distinct ground-floor treatments or story-dependent window styles, remains beyond the current scope.

\subsection{Future Work}
ShellMaker currently assumes structured building scaffolds with predefined wall and opening semantics. It relies on existing layout specifications rather than generating structural layouts directly. Integrating learned 3D layout generation models could enable joint synthesis of exterior geometry and opening configurations within a unified framework. Moreover, in this work, interior–exterior consistency refers to structural preservation (footprint, wall boundaries, and openings) together with prompt-driven stylistic matching of the exterior. Extending this to align generated exteriors with specific interior material and furnishing choices through scaffold refinement is a promising direction for future work.

ShellMaker also retrieves wall and roof textures from curated material libraries to ensure architectural realism and physical plausibility. While effective, this retrieval-based approach limits stylistic diversity. Future work could incorporate generative texture synthesis models to support more controllable, stylized, or unconventional architectural appearances.


\section{Conclusions}
We introduced ShellMaker, a language-guided framework for completing building exteriors from structured scaffolds while strictly preserving structural constraints. Given a scaffold and a style prompt, the system combines parametric roof generation, LLM-based prompt refinement, compatibility-aware material retrieval, and geometry-aware assembly to synthesize coherent exteriors. Overall, ShellMaker provides a practical step toward automated generation of complete building assets with consistent interiors and structurally valid exteriors.

\section{Acknowledgment}
This work is partially supported by NSF Grant \#2411273 and NSF Grant \#2107096.



%
%
\clearpage
\bibliographystyle{splncs04}
\bibliography{main}

\clearpage
\appendix
\section*{ShellMaker Supplementary Material}


\subsection*{Wall and Roof Textures}
We curate a diverse dataset of physically-based rendering (PBR) ready textures for building facades and rooftops, sourced from Poly Haven and CC0 Textures. Wall textures span seven material categories (brick, concrete, plaster, stone, wood, metal, and tile/pavement), while roof textures cover clay/ceramic tiles, slate, metal sheeting, shingle tiles, thatch/reed, and bitumen. Each texture set includes three maps: a diffuse map encoding surface color, a roughness map encoding roughness as a linear grayscale value (0 = mirror-smooth, 1 = fully diffuse), and an OpenGL-convention normal map. For reference, we include 50 representative samples each for wall and roof textures.

\begin{figure}[H]
\centering

\begin{subfigure}{0.48\linewidth}
    \includegraphics[width=\linewidth]{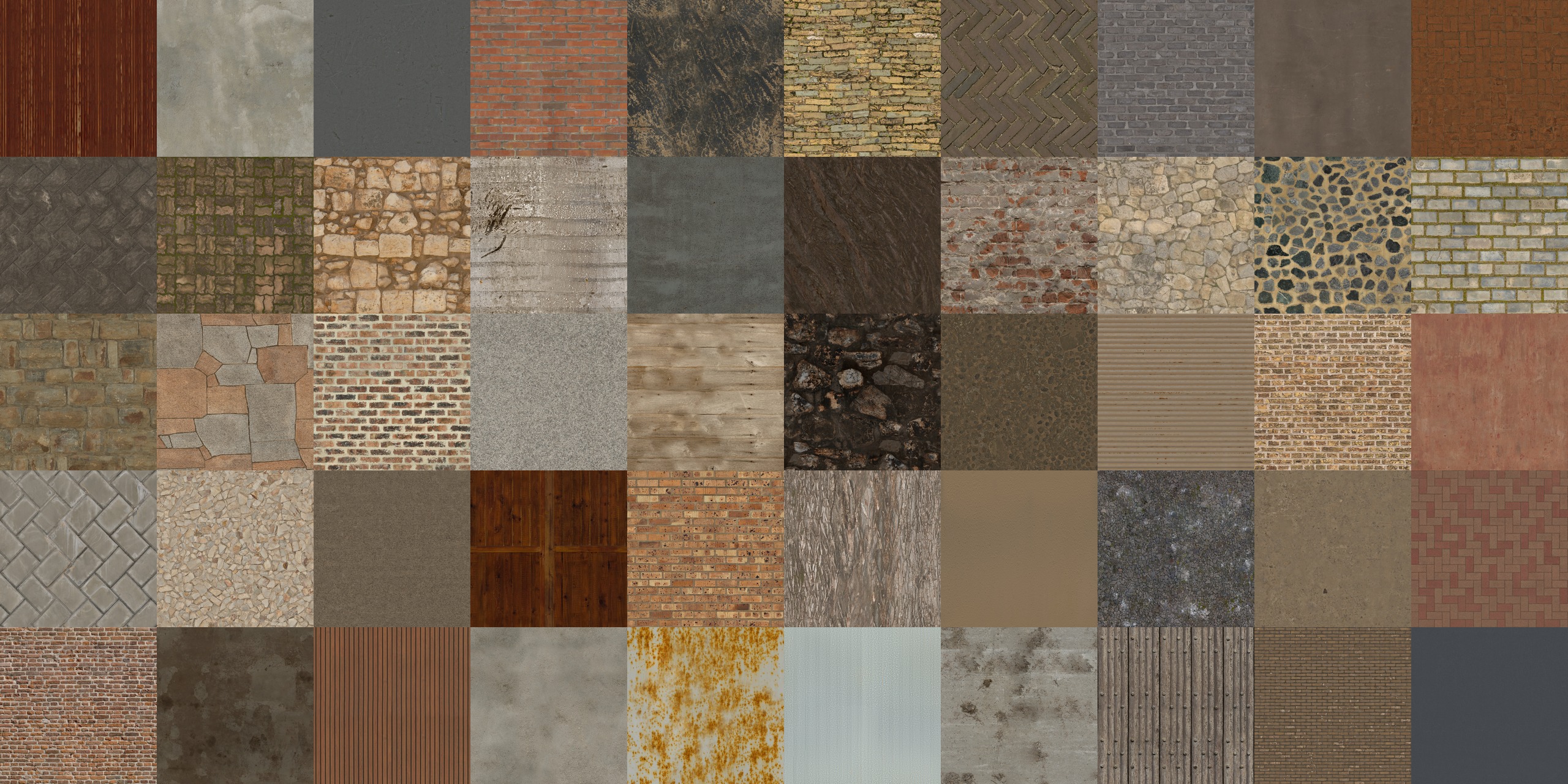}
    \caption{Wall Diffuse}
    \label{fig:wall_diffuse}
\end{subfigure}
\hfill
\begin{subfigure}{0.48\linewidth}
    \includegraphics[width=\linewidth]{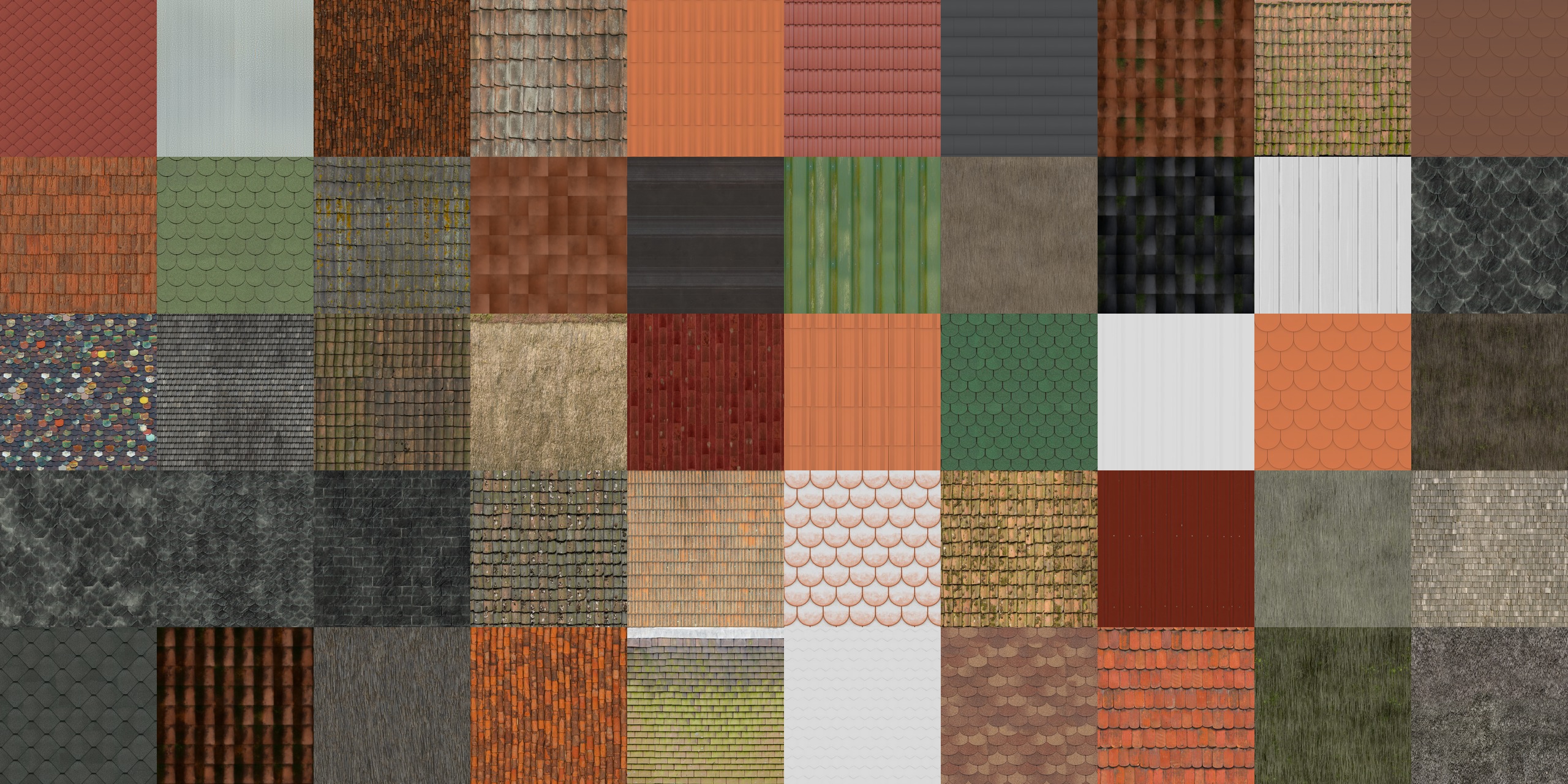}
    \caption{Roof Diffuse}
    \label{fig:roof_diffuse}
\end{subfigure}

\vspace{4pt}

\begin{subfigure}{0.48\linewidth}
    \includegraphics[width=\linewidth]{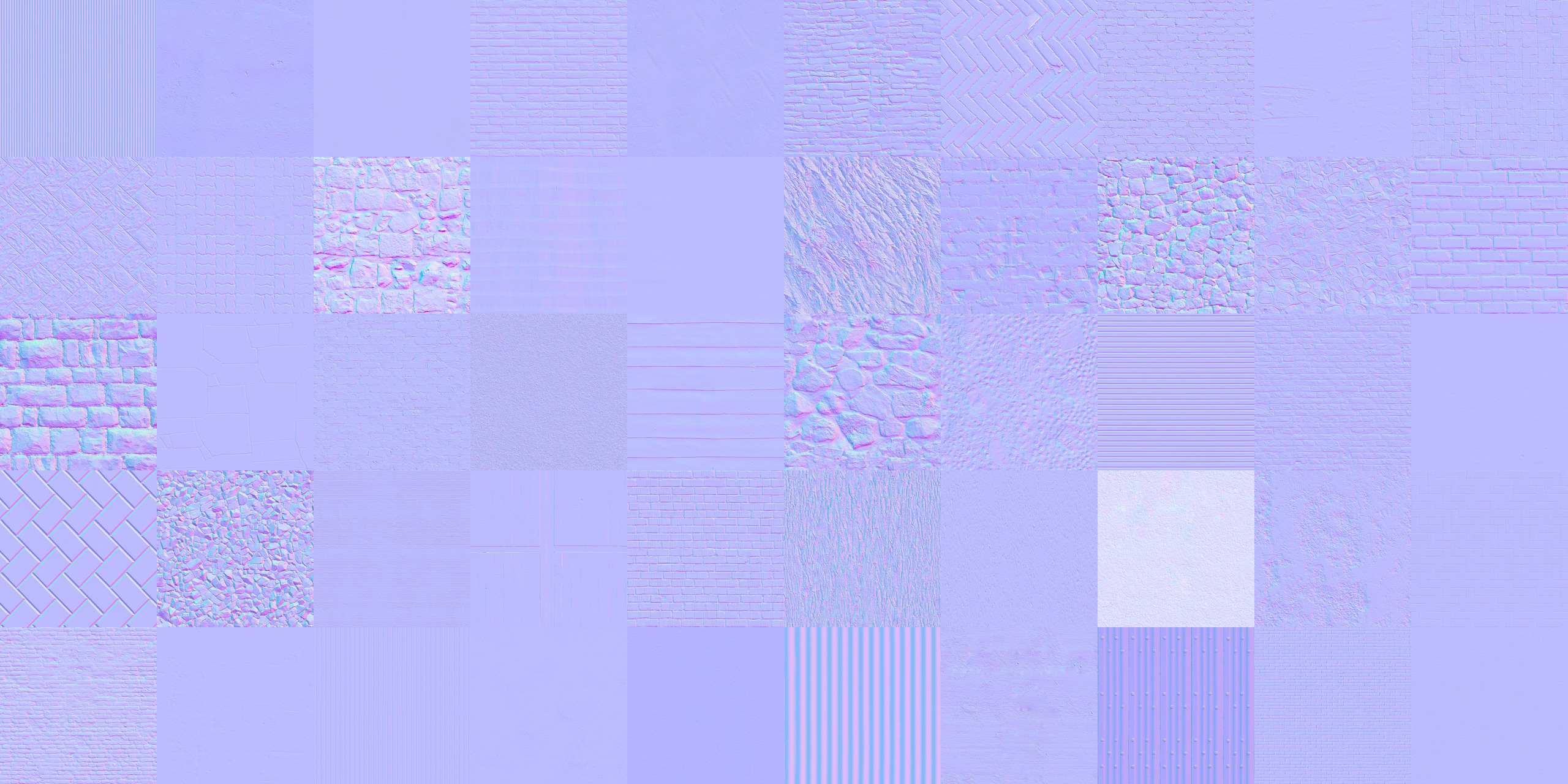}
    \caption{Wall Normal}
    \label{fig:wall_normal}
\end{subfigure}
\hfill
\begin{subfigure}{0.48\linewidth}
    \includegraphics[width=\linewidth]{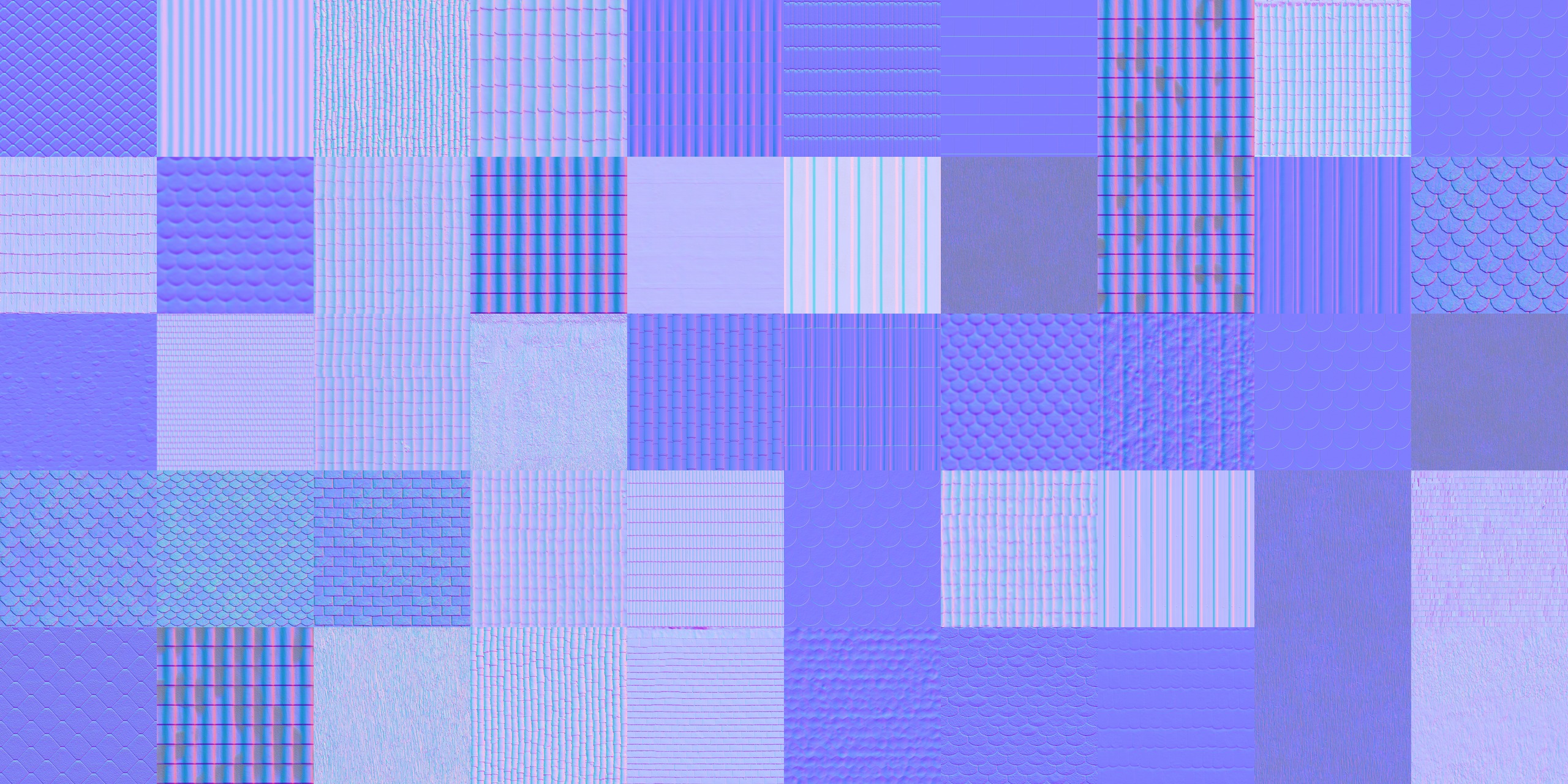}
    \caption{Roof Normal}
    \label{fig:roof_normal}
\end{subfigure}

\vspace{4pt}

\begin{subfigure}{0.48\linewidth}
    \includegraphics[width=\linewidth]{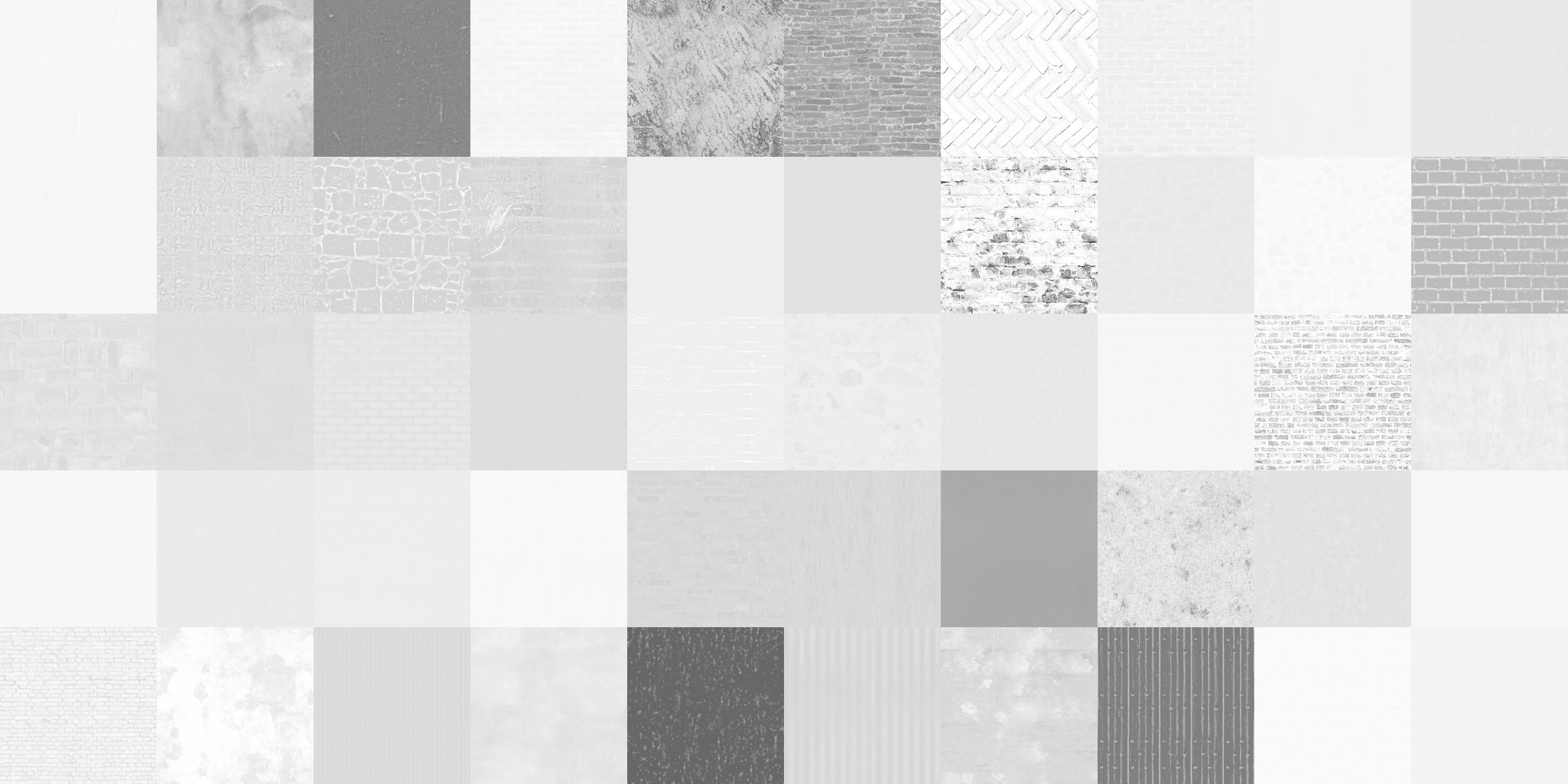}
    \caption{Wall Roughness}
    \label{fig:wall_roughness}
\end{subfigure}
\hfill
\begin{subfigure}{0.48\linewidth}
    \includegraphics[width=\linewidth]{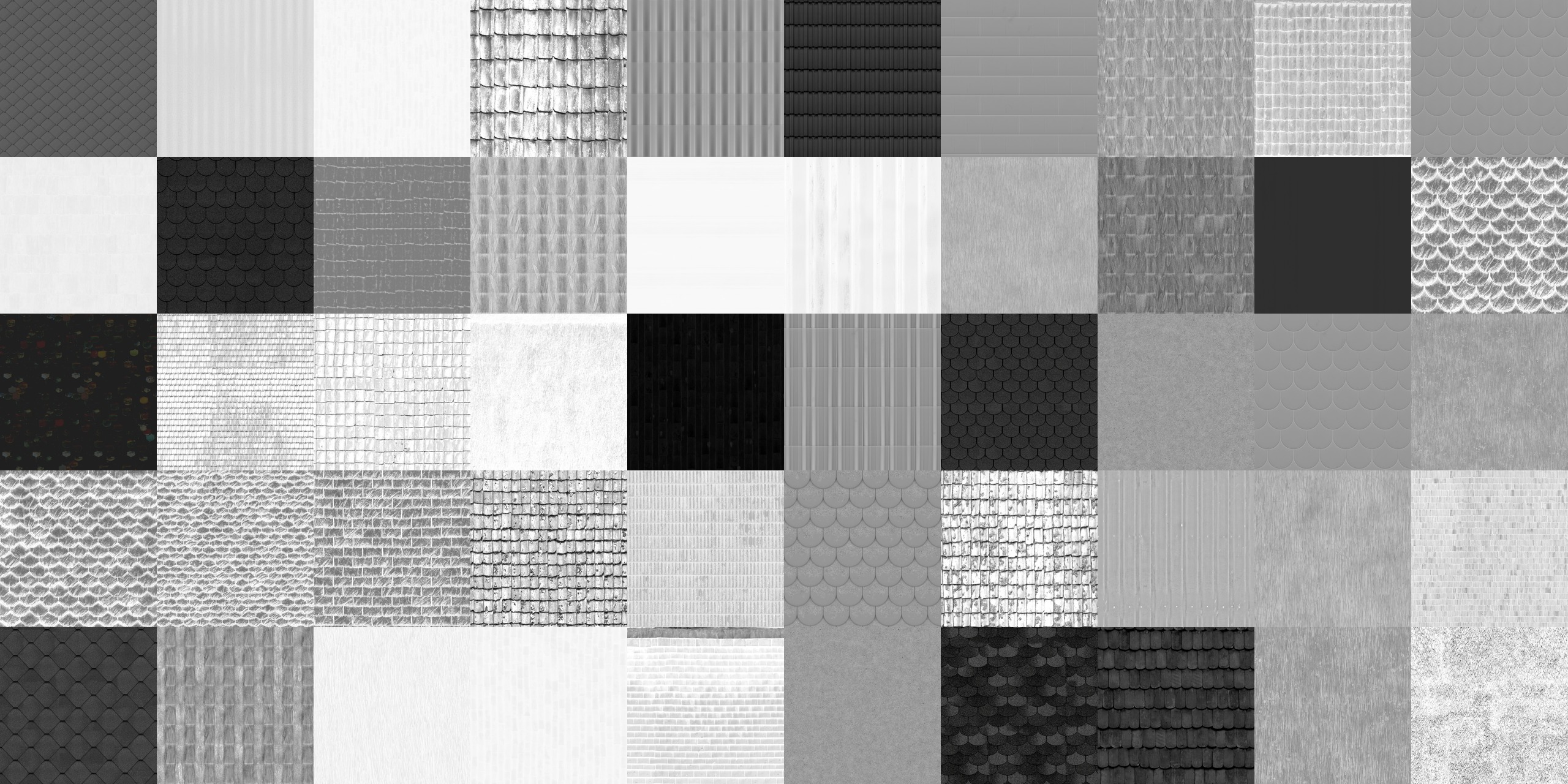}
    \caption{Roof Roughness}
    \label{fig:roof_roughness}
\end{subfigure}

\caption{Representative samples from the curated PBR texture library used by ShellMaker. 
Left column shows facade materials and right column shows roof materials. 
Each material includes diffuse, normal, and roughness maps.}
\label{fig:texture_library}

\end{figure}

\subsection*{Runtime Analysis}
We report wall-clock timings averaged over 500 fully automatic end-to-end runs across diverse architectural styles and floor plans. The mean runtime per scene is 575.71\,s (\(\sim\)9.6\,min). Runtime is dominated by neural asset synthesis during the generation stage.
\begin{itemize}
    \item \textbf{Generation stage (493.23\,s, 85.7\%):} Trellis-2 door/window synthesis 371.18\,s (75.2\%); procedural roof construction and ornaments 121.96\,s (24.7\%); other steps (scaffold parsing, texture retrieval) \(<0.2\) s.
    
    \item \textbf{Assembly stage (82.48\,s):} GLB serialization 51.60\,s (62.6\%); loading/placing Trellis assets 17.77\,s (21.5\%); wall construction and boundary computation 8.81\,s (10.7\%); remaining procedural steps \(<4\) s.
    
    \item \textbf{Primary bottleneck:} neural asset synthesis, which suggests batched or parallelized inference across openings will greatly reduce generation time.
\end{itemize}

\subsection*{Fly-through Visualization}
To better illustrate the spatial coherence of the generated assets, we include a fly-through video of a representative building produced by ShellMaker. The video showcases both the exterior facade and the interior layout populated with furniture, highlighting the consistency between the generated exterior shell and the interior scene. The camera trajectory moves from street-level exterior views into interior rooms, providing a continuous view of the complete building asset.

\subsection*{Failure Case}
\begin{figure}[H]
  \centering
  \begin{subfigure}{0.45\linewidth}
    \includegraphics[width=\linewidth]{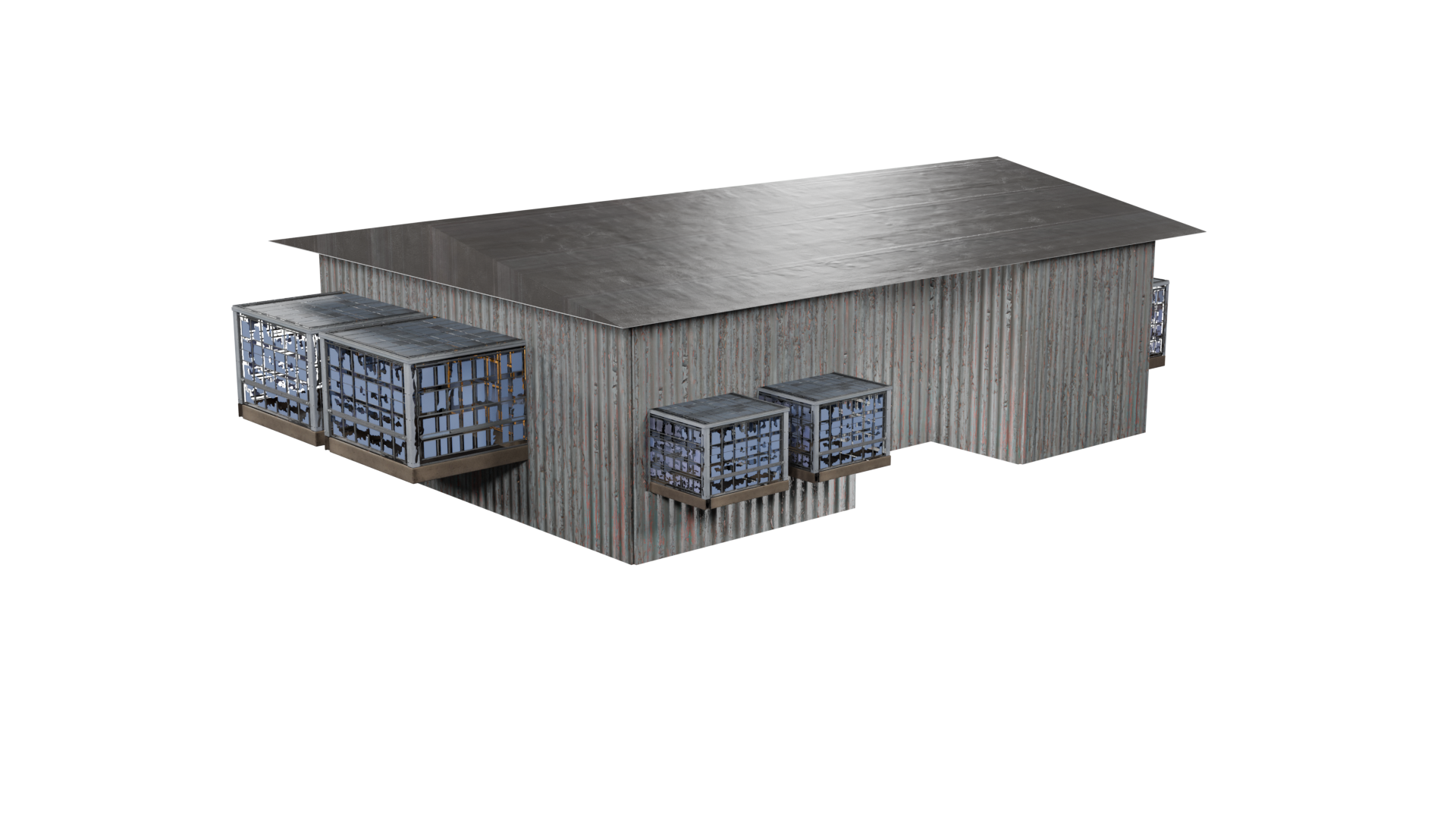}
    \caption{Degenerate facade elements produced by Trellis-2, resulting in malformed window structures.}
    \label{fig:failure-a}
  \end{subfigure}
  \hfill
  \begin{subfigure}{0.45\linewidth}
    \includegraphics[width=\linewidth]{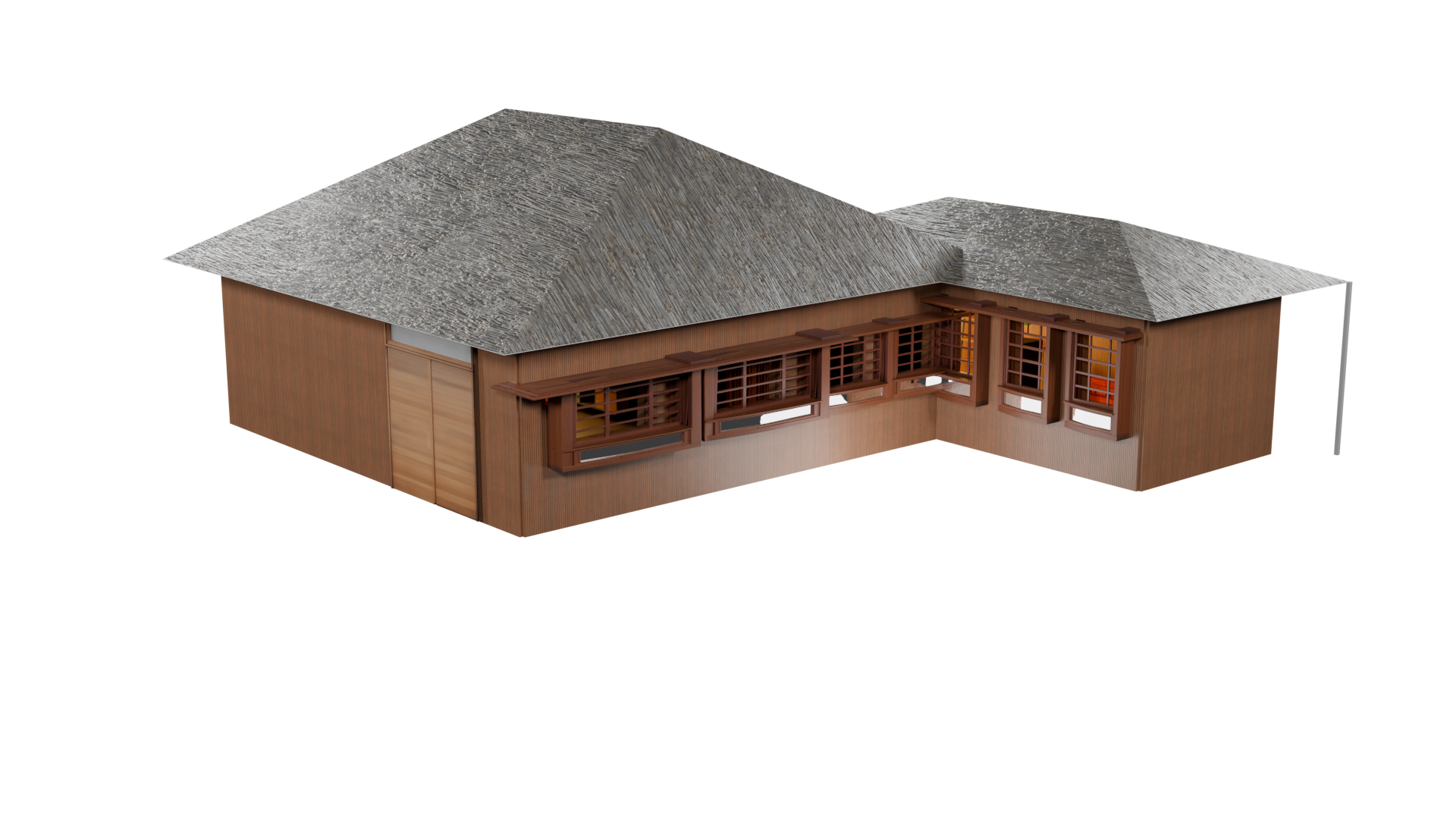}
    \caption{Intersecting facade components caused by dense placement of architectural elements along the same wall segments.}
    \label{fig:failure-b}
  \end{subfigure}
  \caption{Illustrative Failure Cases of ShellMaker.}
  \label{fig:failure}
\end{figure}

Despite generally producing coherent exteriors, ShellMaker can occasionally generate invalid architectural components during the part generation stage (e.g., $<$5\% of the automatic runs). In some cases, the generator produces degenerate facade elements such as malformed or structurally implausible windows (Fig. \ref{fig:failure-a}). In other cases, multiple facade components placed in close proximity may intersect due to imperfect spatial reasoning in the placement stage (Fig. \ref{fig:failure-b}). These issues arise from limitations of the underlying generative models and the absence of strict geometric collision checks in the current pipeline.

\subsection*{Urban Block-Scale Generation}
\begin{figure}[H]
\centering

\begin{subfigure}{\linewidth}
    \includegraphics[width=\linewidth]{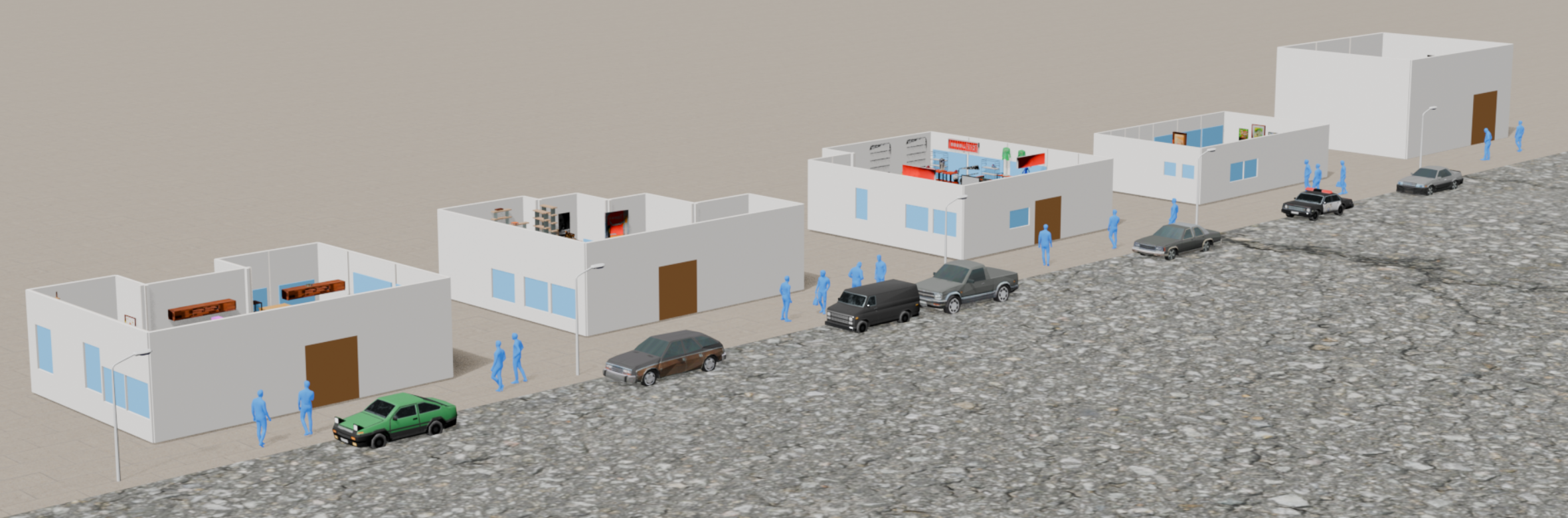}
    \label{fig:street_scene_bbox}
\end{subfigure}
\hfill

\vspace{4pt}

\begin{subfigure}{\linewidth}
    \includegraphics[width=\linewidth]{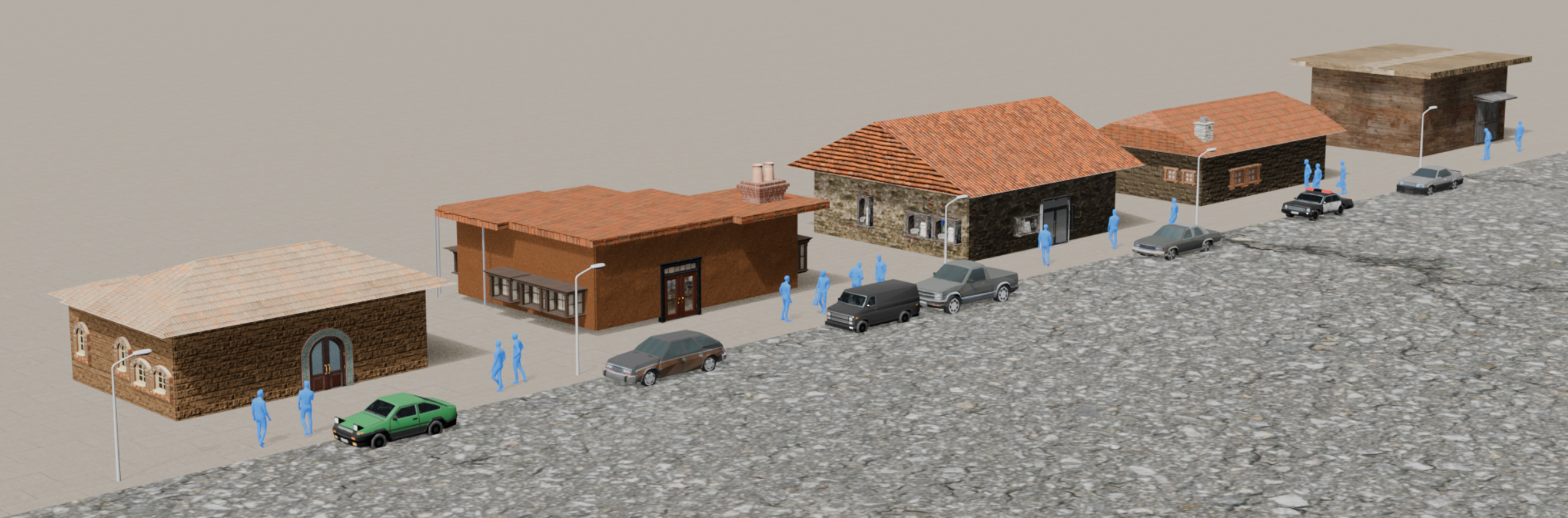}
    \label{fig:street_scene_full}
\end{subfigure}

\caption{\textbf{Urban block-scale generation.} ShellMaker can be applied independently to multiple buildings to synthesize coherent streetscapes. (top) Raw building scaffolds produced by the Holodeck indoor scene generator, showing interior layouts and structural shells. (bottom) Exterior meshes generated by ShellMaker from the same scaffolds, with roofs, materials, and facade details added while preserving the original building footprints.}
\label{fig:urban_block}

\end{figure}

Although ShellMaker operates on individual building scaffolds, it can be applied repeatedly across multiple structures to synthesize coherent urban streetscapes. Fig.~\ref{fig:urban_block} illustrates an example where a row of buildings generated by an indoor scene pipeline (Holodeck) is processed independently by ShellMaker to produce fully textured exterior meshes. Despite being generated per-building, the resulting block exhibits consistent architectural styling and material coherence, demonstrating that ShellMaker can scale naturally to neighborhood- or block-level scenes without modifying the underlying pipeline.

\subsection*{Additional Implementation Details}
\label{sec:additional_impl}

We provide formulations for the color and frequency compatibility signals and list the hyperparameters used in all experiments.

\paragraph{Color Compatibility.}
$C^{\text{color}}$ combines perceptual color distance and saturation difference between the two diffuse maps:
\begin{equation}
C^{\text{color}}_{ij} =
0.7 \left( 1 - \frac{\lVert \Delta \mathbf{LAB} \rVert}{255\sqrt{3}} \right)
+ 0.3 \left( 1 - \lvert \Delta s \rvert \right),
\end{equation}
where $\lVert \Delta \mathbf{LAB} \rVert$ is the mean-color distance in CIELAB
(normalized by the LAB-cube diagonal $255\sqrt{3}$) and $\lvert \Delta s \rvert
\in [0,1]$ is the absolute saturation difference, giving
$C^{\text{color}}_{ij} \in [0,1]$.

\paragraph{Frequency Compatibility.}
For each diffuse map we take a 2D FFT of its luminance image and radially
average the power spectrum into a vector $\mathbf{P} \in \mathbb{R}^{R}$.
$C^{\text{freq}}$ is the cosine similarity of the wall and roof profiles:
\begin{equation}
C^{\text{freq}}_{ij} =
\frac{\mathbf{P}_w \cdot \mathbf{P}_r}
{\lVert \mathbf{P}_w \rVert \, \lVert \mathbf{P}_r \rVert} \in [0,1],
\end{equation}
which lies in $[0,1]$ since the spectrum entries are non-negative.

\paragraph{Retrieval Hyperparameters.}
We shortlist the top $K = 32$ wall and top $K = 32$ roof textures ($K^2$
candidate pairs), retain the top $M = 8$ scored pairs, and sample the final
pair via a softmax with temperature $\tau = 0.12$. Compatibility weights are
$\lambda = (0.45, 0.25, 0.15, 0.15)$ for
$(C^{\text{mat}}, C^{\text{color}}, C^{\text{freq}}, C^{\text{clip}})$ and
pair-scoring weights are $(\alpha, \beta, \gamma) = (0.40, 0.30, 0.30)$ for
$(s_w, s_r, C_{ij})$. All values are fixed across experiments.

\subsection*{Test Prompt Suite}

We define three categories of text prompts used to evaluate the procedural
generation pipeline. \textbf{Style-only} prompts specify architectural
character without material constraints. \textbf{Material-only} prompts
specify wall and roof finishes without any style descriptor.
\textbf{Mixed} prompts combine a style descriptor with explicit material
clauses.

\subsubsection*{Category 1: Style-Only (20 prompts)}

\begin{enumerate}[label=\textbf{S\arabic*.}, leftmargin=*, nosep]
  \item a Victorian colonial mansion
  \item a rustic farmhouse
  \item a modern minimalist residential building
  \item a Mediterranean villa
  \item a Moorish North African riad
  \item a Japanese traditional house
  \item an industrial warehouse conversion
  \item a Scandinavian timber house
  \item a Proven\c{c}al French country farmhouse
  \item a Gothic revival townhouse
  \item an Art Deco apartment building
  \item a Baroque palace
  \item a Tudor manor house
  \item a Bauhaus residential block
  \item a Spanish Colonial hacienda
  \item a Cape Cod cottage
  \item a Brutalist civic building
  \item an Ottoman-style pavilion
  \item a Chinese traditional courtyard house
  \item a Craftsman bungalow
\end{enumerate}

\subsubsection*{Category 2: Material-Only (20 prompts)}

\begin{enumerate}[label=\textbf{M\arabic*.}, leftmargin=*, nosep]
  \item a building with red brick walls and a grey slate roof
  \item a building with white stucco walls and a terracotta tile roof
  \item a building with rough grey stone walls and a dark wood shingle roof
  \item a building with timber-frame walls and a thatched roof
  \item a building with sandstone walls and a flat clay tile roof
  \item a building with weathered wood walls and a green copper roof
  \item a building with white-washed plaster walls and an orange clay tile roof
  \item a building with dark timber walls and a black standing-seam metal roof
  \item a building with exposed concrete walls and a flat gravel roof
  \item a building with yellow limestone walls and a red terracotta tile roof
  \item a building with flint stone walls and a grey slate roof
  \item a building with blue-grey granite walls and a dark slate roof
  \item a building with adobe mud walls and a flat earthen roof
  \item a building with pale limestone walls and a blue glazed ceramic tile roof
  \item a building with rough-hewn stone walls and a corrugated metal roof
  \item a building with white marble cladding and a glass and steel roof
  \item a building with brown brick walls and a green clay tile roof
  \item a building with cedar shingle walls and a wooden shake roof
  \item a building with ochre-painted plaster walls and a terracotta pan tile roof
  \item a building with polished concrete walls and a standing-seam zinc roof
\end{enumerate}

\subsubsection*{Category 3: Mixed (20 prompts)}

\begin{enumerate}[label=\textbf{X\arabic*.}, leftmargin=*, nosep]
  \item Victorian colonial style with red brick walls and a grey slate roof
  \item rustic farmhouse style with rough stone walls and a thatched roof
  \item modern style with white concrete walls and a flat glass roof
  \item Mediterranean style with white stucco walls and a terracotta tile roof
  \item Moorish North African style with white-washed plaster walls and a green glazed tile roof
  \item Japanese style with dark timber walls and a grey ceramic tile roof
  \item industrial style with exposed red brick walls and a corrugated metal roof
  \item Scandinavian style with pale birch timber walls and a dark green metal roof
  \item Proven\c{c}al French style with ochre limestone walls and a terracotta pan tile roof
  \item Gothic style with grey limestone walls and a dark slate roof
  \item Art Deco style with cream limestone cladding and a flat copper parapet roof
  \item Baroque style with pale sandstone walls and a green copper mansard roof
  \item Tudor style with white-washed plaster and dark oak timber-frame walls and a dark shingle roof
  \item Bauhaus style with white render walls and a flat black tar roof
  \item Spanish Colonial style with white stucco walls and a red clay tile roof
  \item Cape Cod style with white-painted clapboard walls and a grey wood shingle roof
  \item Brutalist style with board-formed raw concrete walls and a flat gravel roof
  \item Ottoman style with pale sandstone walls and a lead-grey domed roof
  \item Chinese traditional style with white plaster walls and a glazed green ceramic tile roof
  \item Craftsman style with brown river-stone base walls, cedar shingle upper walls, and a dark wood shake roof
\end{enumerate}

\subsection*{Part-Aware LLM Refinement Prompt Bundle}

As explained in Sec. 3.3, we employ an instruction-tuned LLM to automatically refine the input free-form prompt into a set of structured text-to-3D prompts. In the following, we list the prompts used in the Part-aware Prompt Refinement Stage to help with system reproducibility.
\subsubsection*{Wall}
\begin{itemize}
    \item \textbf{System Message:} \\
    You recommend a single exterior wall material and texture for a building based on its architectural style. Choose a material that is authentic to the described style (e.g. exposed red brick for industrial, white stucco plaster for Mediterranean, half-timber with wattle-and-daub for Tudor, raw concrete for Brutalist, rough-cut limestone blocks for Gothic, painted wood clapboard for Colonial, cedar shingle siding for Craftsman, smooth render for Modern). Return ONLY a JSON object with a single key "texture" whose value is a concise material description (3-8 words).

    \item \textbf{User Message:} \\
    Building info: \{style\_prompt\}
    Recommend one exterior wall material/texture for this building.
    The description should be a concise PBR-style texture label (e.g. "aged red brick wall", "smooth white stucco plaster", "rough-cut grey limestone blocks").
\end{itemize}

\subsubsection*{Roof}
\begin{itemize}
    \item \textbf{System Message:} \\
    You recommend a single roofing material and texture for a building based on its architectural style. Choose a material that is authentic to the described style (e.g. terracotta clay tiles for Mediterranean, dark grey slate for Victorian, thatched reed for cottage, standing-seam metal for Modern, cedar wood shakes for Craftsman, asphalt shingles for suburban Colonial, corrugated iron for industrial). Return ONLY a JSON object with a single key "texture" whose value is a concise material description (3-8 words).

    \item \textbf{User Message:} \\
    Building info: \{style\_prompt\}
    Recommend one roofing material/texture for this building.
    The description should be a concise PBR-style texture label (e.g. "terracotta clay roof tiles", "dark grey natural slate", "weathered cedar wood shakes").
\end{itemize}

\subsubsection*{Door Entrance}
\begin{itemize}
    \item \textbf{System Message:} \\
    You generate concise text prompts for 3D building entrance models. Each prompt must describe a single isolated architectural element — the entrance only, with NO surrounding walls, ground plane, or environment. Clean background that is clearly distinguishable from the object, no surroundings. Each prompt must include a front door as the main object and may include surrounding details authentic to the architectural style (e.g. pointed arches and iron hardware for Gothic, flush frameless entries for Modern, wrought iron and tile accents for Mediterranean, porch stoops for farmhouse, etc.).
    \item \textbf{User Message:} \\
    Building info: \{style\_prompt\}
    Generate one text-to-3D prompt per key.
    Constraints: 
    \begin{enumerate}
        \item Each prompt MUST describe an isolated architectural element with clean background and no surroundings.
        \item Do NOT include walls, buildings, ground, or any environment context.
        \item Choose entrance details that are authentic to the architectural style described above. Do NOT default to generic porch/awning/handrail elements unless they suit the style.
    \end{enumerate}
\end{itemize}

\subsubsection*{Window}
\begin{itemize}
    \item \textbf{System Message:} \\
    You generate concise text prompts for 3D window models. Each prompt must describe a single isolated architectural element — the window only, with NO surrounding walls, ground plane, or environment. Clean background, no surroundings. Each prompt must include the window frame and glass as the main object and may include details authentic to the architectural style (e.g. pointed arches and tracery for Gothic, leaded diamond panes for Tudor, iron grilles and shutters for Mediterranean, clean frameless glazing for Modern, etc.). [If material is glass: The window glass must be transparent and clear.] 
    \item \textbf{User Message:} \\
    Building info: \{style\_prompt\}
    Generate one text-to-3D prompt per key.
    Constraints: 
    \begin{enumerate}
        \item Each prompt MUST describe an isolated architectural element with clean background and no surroundings.
        \item Do NOT include walls, buildings, ground, or any environment context.
        \item Choose window details that are authentic to the architectural style described above. Do NOT default to generic muntins/shutters/sill unless they suit the style. [If material is glass: - Glass must be transparent and clear.]
    \end{enumerate}
\end{itemize}

\subsubsection*{Roof Ornament}
\begin{itemize}
    \item \textbf{System Message:}\par
    You generate concise text prompts for 3D roof ornament models. Allowed ornament categories: \{cat\_descriptions\}. Each prompt must describe a single isolated architectural element --- the ornament only, with NO roof surface, ground, walls, or surrounding environment. Clean background, no surroundings. The ornament's material, shape, and detailing must be authentic to the building's architectural style (e.g.\ crocketed stone pinnacles for Gothic, twisted brick chimneys for Tudor, terracotta elements for Mediterranean, clean metal for Modern/Scandinavian, weathered iron for rustic, etc.). The color palette should complement the roof material. Use a VARIETY of the allowed categories across the prompts --- do NOT repeat the same category for every prompt. Return ONLY a JSON object with a single ``prompts'' key whose value is an array of strings. Example: \{"prompts": ["prompt one", "prompt two"]\}. No markdown, no prose.

    \item \textbf{User Message:}\par
    Building info: \{style\_prompt\}\par
    Roof type: \{roof\_type\}\par
    {[}Roof material/texture: \{roof\_texture\_hint\}{]} (appended only when available)\par
    Generate exactly \{count\} distinct roof ornament prompts, choosing from these categories: \{ornament\_categories\}. Each prompt MUST describe a single isolated ornament as an architectural element with clean background and no surroundings --- no roof surface, no walls, no environment. Choose ornament details that are authentic to the architectural style described above. Do NOT default to generic ornament forms --- match the style's characteristic materials and shapes. The color palette should complement the roof material.
\end{itemize}
\end{document}